\documentclass[11pt, a4paper, onecolumn, thu]{thuc3i}

\usepackage[authoryear, sort&compress, square]{natbib}
\usepackage{fontawesome5}

\usepackage{xspace}
\usepackage[table]{xcolor}

\usepackage[skip=6pt]{caption}
\definecolor{lightblue}{RGB}{220,235,250}

\usepackage{lipsum}
\usepackage{float}

\usepackage[most,skins,theorems]{tcolorbox}
\tcbset{
  takeawaysbox/.style={
    title=Takeaways,
    colback=lightblue!80,
    colframe=black,
    fonttitle=\bfseries\small,
    coltitle=white,
    colbacktitle=black,
    enhanced,
    attach boxed title to top left={xshift=2.5mm,yshift=-2.5mm},
    boxed title style={rounded corners, size=small, colframe=black, colback=black},
    width=\linewidth,
    arc=3.5mm
  }
}

\usepackage{graphicx}
\usepackage{algorithm}
\usepackage{algpseudocode}

\usepackage{subcaption}
\usepackage{multirow}
\usepackage{wrapfig}

\usepackage[dvipsnames]{xcolor}
\usepackage{hyperref}
\usepackage{makecell}
\usepackage{listings}
\lstnewenvironment{normalcode}{
  \lstset{
    language=Python,
    basicstyle=\ttfamily\small,
    backgroundcolor=\color{white},
    breaklines=true,
    showstringspaces=false,
    frame=none,
    xleftmargin=0.5em,
    aboveskip=1pt,
    belowskip=1pt
  }
}{}

\lstnewenvironment{diffaddcode}{
  \lstset{
    language=Python,
    basicstyle=\ttfamily\small,
    backgroundcolor=\color{green!15},
    breaklines=true,
    showstringspaces=false,
    frame=none,
    xleftmargin=0.5em,
    aboveskip=1pt,
    belowskip=1pt
  }
}{}

\lstnewenvironment{diffdelcode}{
  \lstset{
    language=Python,
    basicstyle=\ttfamily\small,
    backgroundcolor=\color{red!15},
    breaklines=true,
    showstringspaces=false,
    frame=none,
    xleftmargin=0.5em,
    aboveskip=1pt,
    belowskip=1pt
  }
}{}
\DeclareCaptionType{listing}[Listing][List of Listings]

\newcommand{\method}{\textbf{SimpleVLA-RL}\xspace}

\NewDocumentCommand{\yuxin}{ mO{} }{\textcolor{magenta}{\textsuperscript{yuxin}\textbf{\small[#1]}}}

\NewDocumentCommand{\haozhan}{ mO{} }{\textcolor{red}{\textsuperscript{haozhan}\textbf{\small[#1]}}}

\NewDocumentCommand{\todo}{ mO{} }{\textcolor{blue}{\textsuperscript{Todo}\textbf{[#1]}}}

\NewDocumentCommand{\ning}{ mO{} }{\textcolor{blue}{\textsuperscript{ning}\textbf{\small[#1]}}}

\definecolor{darkred}{RGB}{200, 0, 0}

\setheadertext{SimpleVLA-RL Technical Report}

\title{SimpleVLA-RL: Scaling VLA Training via Reinforcement Learning}

\reportnumber{} %

\renewcommand{\today}{2025-09-12}

\author[*,1]{Haozhan Li}
\author[*,1]{Yuxin Zuo}
\author[*,1]{Jiale Yu}
\author[*,3]{Yuhao Zhang}
\author[3]{Zhaohui Yang}
\author[1]{Kaiyan Zhang}
\author[1]{Xuekai Zhu}
\author[4]{Yuchen Zhang}
\author[5]{Tianxing Chen}
\author[2]{Ganqu Cui}
\author[3]{Dehui Wang}
\author[3]{Dingxiang Luo}
\author[3]{Yuchen Fan}
\author[1]{Youbang Sun}
\author[2]{Jia Zeng}
\author[2]{Jiangmiao Pang}
\author[4]{Shanghang Zhang}
\author[1]{Yu Wang}
\author[2,3]{Yao Mu}
\author[$\dagger$,1,2]{Bowen Zhou}
\author[$\dagger$,1,2]{Ning Ding}

\affil[1]{\thepa{}{}}
\affil[2]{Shanghai AI Lab}
\affil[3]{Shanghai Jiao Tong University}
\affil[4]{Peking University}
\affil[5]{The University of Hong Kong}

\role[*]{Equal Contributions}
\role[\dagger]{Corresponding Authors}

\resource{\faEnvelope[regular]}{zhan72426@gmail.com, dingning@mail.tsinghua.edu.cn}
\resource{\faGithub}{\href{https://github.com/PRIME-RL/SimpleVLA-RL}{PRIME-RL/SimpleVLA-RL}}

\begin{abstract}
Vision-Language-Action (VLA) models have recently emerged as a powerful paradigm for robotic manipulation. Despite substantial progress enabled by large-scale pretraining and supervised fine-tuning (SFT), these models face two fundamental challenges:
(i) the scarcity and high cost of large-scale human-operated robotic trajectories required for SFT scaling,
and (ii) limited generalization to tasks involving distribution shift.
Recent breakthroughs in Large Reasoning Models (LRMs) demonstrate that reinforcement learning (RL) can dramatically enhance step-by-step reasoning capabilities, raising a natural question: \textbf{\textit{Can RL similarly improve the long-horizon step-by-step action planning of VLA?}}
In this work, we introduce \method, an efficient RL framework tailored for VLA models.
Building upon veRL, we introduce VLA-specific trajectory sampling, scalable parallelization, multi-environment rendering, and optimized loss computation.
When applied to OpenVLA-OFT, \method achieves SoTA performance on LIBERO and even outperforms $\pi_0$ on RoboTwin 1.0\&2.0 with the exploration-enhancing strategies we introduce.
\method not only reduces dependence on large-scale data and enables robust generalization, but also remarkably surpasses SFT in real-world tasks.
Moreover, we identify a novel phenomenon ``\textcolor{darkred}{pushcut}'' during RL training, wherein the policy discovers previously unseen patterns beyond those seen in the previous training process.

\end{abstract}

\begin{document}

\maketitle

\begin{figure}[h]
\centering
\includegraphics[width=.99\textwidth]{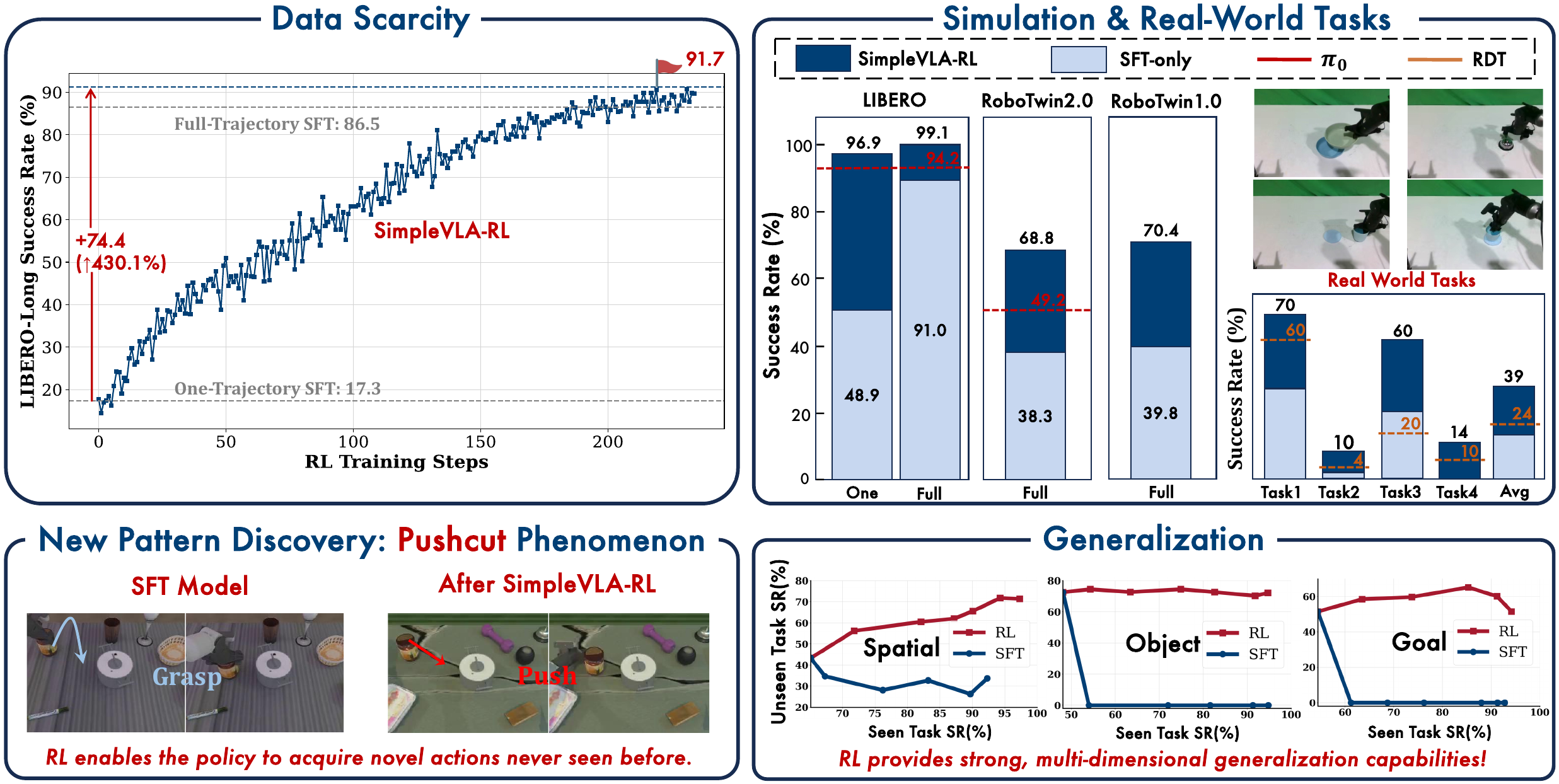}
\caption{
Overview of \method{}.
\method is an efficient RL framework for VLA that improves long-horizon planning under data scarcity, outperforms SFT in simulation and real-world tasks, reveals a ``\textcolor{darkred}{pushcut}'' new-action phenomenon, and strengthens spatial/object/goal generalization.
}
\label{fig:teaser}
\end{figure}

\newpage
\begingroup
\setlength{\baselineskip}{\baselineskip}
\tableofcontents
\endgroup
\newpage

\section{Introduction}

Vision-Language-Action (VLA) models have emerged as a promising paradigm for enabling robots to solve diverse and challenging manipulation tasks in physical environments~\citep{firoozi2025foundation}.
While these models demonstrate considerable potential, their development entails substantial complexity, as they necessitate the unification of visual perception, language understanding, and action generation within a unified framework~\citep{kim2024openvla,zhong2025survey}.
To this end, previous works have typically adopted a two-stage training strategy: first conducting large-scale pretraining on abundant multimodal data with human manipulation videos~\citep{sapkota2025vision}, image-text pairs, and heterogeneous robot datasets~\citep{o2024open}, followed by supervised fine-tuning (SFT) on extensive high-quality robot trajectories to enhance task-specific capabilities.

This paradigm has achieved substantial progress, particularly through large-scale pretraining.
However, as model training scales further, especially when requiring scaled SFT for downstream task optimization, several critical challenges have emerged:
\textbf{1) Data Scarcity:} Scaling SFT necessitates a substantially larger amount of human-operated robot trajectories, which remain both scarce and prohibitively expensive, posing significant challenges to scalability.
The collection of robotic trajectories demands meticulously designed experimental scenarios, diverse manipulation objects, and skilled operators, which severely constrains both data scale and diversity~\citep{mandlekar2021matters,bu2025agibot,gao2024efficient,team2025gemini}.
\textbf{2) Poor Generalization:} Generalization remains a crucial bottleneck for VLA, particularly on compositional, long-horizon, or real-world tasks involving distribution shift.
This challenge is particularly evident because the current SFT of VLAs typically relies on limited, scene- and task-specific data.
Consequently, when VLA models encounter unseen tasks, environments, or objects, their performance inevitably degrades~\citep{liu2025can}.

Recent advances in Large Reasoning Models (LRMs), such as DeepSeek-R1~\citep{guo2025deepseek}, have demonstrated that reinforcement learning (RL) can drive remarkable progress even when relying solely on outcome rewards.
These findings highlight the crucial role of RL in enhancing models’ ability to perform step-by-step chain-of-thought (CoT) reasoning~\citep{team2025kimi,zeng2025simplerl,yang2025qwen3}.
This naturally raises a question:
\textbf{\textit{Can RL also strengthen VLA models’ capacity to generate accurate actions step by step?}}
This insight motivates us to explore reproducing the success of DeepSeek-R1 in VLA models, which may also help overcome the aforementioned two challenges of SFT.
However, applying RL to VLAs actually presents several unique challenges.
Traditional RL methods in robotic tasks typically rely on hand-crafted process rewards, which severely limit scalability~\citep{ibarz2021train,kroemer2021review,ma2023eureka}.
Moreover, unlike the rollout of LLMs, VLAs require multi-round interactions with the environment.
It is not only slower but also substantially more costly.

We introduce \method, an effective RL framework for VLA models.
Building upon Volcano Engine Reinforcement Learning for LLMs (veRL)\footnote{\url{https://github.com/volcengine/verl}}~\citep{sheng2024hybridflow}, a general-purpose RL framework for LLMs, we enable end-to-end online rule-based RL for VLA models through the implementation of VLA-specific interactive trajectory sampling and loss computation.
To further support scalable RL for VLA models, we extend veRL with parallel multi-environment rendering for faster sampling, and adapt it into an integrated training–inference–rendering framework.
Surprisingly, we observe that the policy discovers previously unseen patterns during RL training, extending beyond those encountered in supervised data.
We refer to this novel phenomenon as ``\textcolor{darkred}{pushcut}''.
Our main contributions include:

\begin{itemize}
\item \textbf{Efficient online RL framework for VLA:} We develop an efficient end-to-end VLA online RL framework based on veRL that enables stable, sample-efficient training, optimized for rendering parallelization and distributed training \& inference.
\item \textbf{SoTA performance:} We incorporate exploration-enhancing strategies, yielding consistent performance improvements of 10–15\%. Moreover, \method surpasses multiple SoTA baselines on both LIBERO and RoboTwin 1.0 \& 2.0.
\item \textbf{Data efficiency and generalization:} With only a single demonstration per task, RL boosts LIBERO-Long success rates from 17.1\% to 91.7\%, and significantly outperforms SFT in spatial, object, and task generalization.
\item \textbf{Real-world deployment capability:} Simulation-trained policies transfer effectively to real-world, achieving strong sim-to-real performance gains without requiring any real robot data.
\end{itemize}

\section{Preliminaries}
\label{sec:preliminaries}

To provide an intuitive illustration of the existing gap when extending RL methodologies from LLMs to the VLA domain, we formalize RL for both LLMs and VLA models, presenting their state representations, action spaces, reward functions, and environments in this section.

\subsection{RL Formulation for LLMs}
\textbf{State ($s_t$):} At step $t$, the state $s_t$ comprises the input prompt and previously generated tokens:
\begin{equation}
    s_t = (x_{\text{prompt}}, y_1, y_2, \ldots, y_{t-1}),
\end{equation}
where $x_{\text{prompt}}$ denotes the initial prompt and $y_t$ denotes the $t$-th generated token.

\textbf{Action ($a_t$):} An action corresponds to selecting the next token from the vocabulary $\mathcal{V}$.
At each step, the policy outputs a probability distribution over tokens, and the action token is selected via random sampling. Formally, the action is defined as:
\begin{equation}
a_t = y_t \in \mathcal{V}, \quad \text{where} \quad y_t \sim \pi_\theta(\cdot|s_t) = \mathrm{softmax}\left(f_\theta(s_t) / T\right),
\end{equation}
where $f_\theta(s_t) \in \mathbb{R}^{|\mathcal{V}|}$ represents the LLM logit outputs and $T$ is the temperature parameter controlling the randomness of sampling.

\textbf{Environment:} The environment provides reward signals upon sequence completion. In rule-based settings, binary rewards are assigned based on the correctness.
Alternatively, learned reward models or human feedback systems provide continuous rewards based on criteria such as helpfulness, harmlessness, or task alignment. The reward is computed as follows:
\begin{equation}
r(\tau) = \begin{cases}
1, & \text{if } \tau \text{ satisfies correctness criteria} \\
0, & \text{otherwise}
\end{cases},
\quad\text{or}\quad r(\tau) = R_\phi(\tau) \in [0, 1],
\end{equation}
where $R_\phi$ is a learned reward model and $\tau = (x_{\text{prompt}}, y_1, y_2, \ldots, y_{T_{\text{seq}}})$ represents the complete generated sequence of length $T_{\text{seq}}$.

\textbf{Rollout:} 
Given an input prompt $x_{\text{prompt}}$, the LLM auto-regressively generates a sequence by sampling tokens from $\pi_\theta(y_t|s_t)$ until termination, 
without intermediate environmental feedback. 
With a non-zero temperature $T$, the policy can produce diverse rollouts that explore different solution paths.

\subsection{RL Formulation for VLAs}

\textbf{State ($s_t$):} The state consists of multimodal observations including visual input (RGB images, depth maps, or point clouds), proprioceptive information (joint angles, end-effector pose), and language instructions of the tasks. Formally, the state is defined as:
\begin{equation}
s_t = (o_t^{\text{vis}}, o_t^{\text{prop}}, l_{\text{task}}),
\end{equation}
where $o_t^{\text{vis}}$ is multimodal observations, $o_t^{\text{prop}}$ is proprioceptive information, and $l_{\text{task}}$ is language instructions.

\textbf{Action ($a_t$):} Actions are control commands in the robot action space, typically end-effector deltas or joint angle targets, where $a_t \in \mathbb{R}^d$ (e.g., $d=7$ for 6-DoF pose plus gripper position).
Most VLA policies generate actions through either a diffusion-based action expert or a discrete action tokenizer.
The action is defined as follows:
\begin{equation}
a_t = {Decoder}(h_\theta(s_t)), \quad {Decoder} \in \{\text{Diffusion Expert}, \text{Action Tokenizer}\}, \quad a_t \in \mathbb{R}^d,
\end{equation}
where $h_\theta(s_t)$ represents the hidden state of $s_t$ in the VLA model, and $Decoder$ is the action decoder.

\textbf{Environment:} The environment represents the physical world or simulation where the robot operates. It provides state transitions $s_{t+1}=\text{Env}(s_t, a_t)$ and reward signals:
\begin{equation}
r_t = \alpha \cdot {I}_{\text{success}} + (1-\alpha) \cdot \sum_i w_i \cdot \phi_i(s_t, a_t), \quad \alpha \in [0,1], \quad {I}_{\text{success}} = \begin{cases}
1, & \text{if task success} \\
0, & \text{otherwise}
\end{cases},
\end{equation}
where $\phi_i(s_t, a_t)$ represents process rewards (e.g. distance to goal), $w_i$ are weights, and $\alpha$ balances outcome and process rewards.

\textbf{Rollout:} VLA models generate trajectories through iterative interaction with the environment.
At each timestep, the policy $\pi_\theta$ takes the current state $s_t$ as input and outputs an action chunk $(a_t, a_{t+1}, \ldots, a_{t+k-1})$ of length $k$.
The robot executes these actions sequentially and the environment produces updated states based on physical dynamics.
After execution, the model takes the new state $s_{t+k}$ as input and generates the next action chunk. This process continues until task completion or maximum episode length, producing a complete trajectory $\tau = ((s_0, a_0), (s_1, a_1), \ldots, (s_T, a_T))$ through interactive sampling.

\subsection{Group Relative Policy Optimization}
\label{sec:grpo}
Group Relative Policy Optimization~(GRPO)~\citep{shao2024deepseekmath} is an RL method that eliminates the value function by computing advantages through group-relative normalization.
Given an initial state $s_0$, the behavior policy $\pi_{\theta_{\text{old}}}$ generates $G$ trajectories $\{\tau_i\}_{i=1}^G$.
The GRPO objective employs PPO-style clipping with KL regularization to constrain policy updates:
\begin{equation}
\label{eq:grpo_objective}
\begin{split}
J_{\text{GRPO}}(\theta) = \mathbb{E}_{s_0 \sim \mathcal{D}, \{\tau_i\} \sim \pi_{\theta_{\text{old}}}} \Bigg[ \frac{1}{G} \sum_{i=1}^G \frac{1}{|\tau_i|} \sum_{t=1}^{|\tau_i|} & \min\left( r_{i,t}(\theta) \hat{A}_i, \text{clip}(r_{i,t}(\theta), 1-\epsilon, 1+\epsilon) \hat{A}_i \right) \\
& - \beta D_{\text{KL}}(\pi_\theta || \pi_{\text{ref}}) \Bigg],
\end{split}
\end{equation}
where the importance sampling ratio $r_{i,t}(\theta)$ and the normalized advantage $\hat{A}_i$ are defined as:
\begin{equation}
\label{eq:grpo_components}
r_{i,t}(\theta) = \frac{\pi_\theta(a_{i,t}|s_{i,t})}{\pi_{\theta_{\text{old}}}(a_{i,t}|s_{i,t})}, \quad \hat{A}_i = \frac{R_i - \text{mean}(\{R_i\}_{i=1}^G)}{\text{std}(\{R_i\}_{i=1}^G)}.
\end{equation}
Here $R_i$ denotes the total reward of the $i$-th trajectory, $\epsilon > 0$ is the PPO clipping parameter that limits the policy ratio, and $\beta > 0$ is the coefficient controlling the strength of KL regularization with respect to the reference policy $\pi_{\text{ref}}$.

\section{\method}
\label{sec:method}

\begin{figure}[!tbp]
  \centering
  \includegraphics[width=\textwidth,height=0.33\textheight]{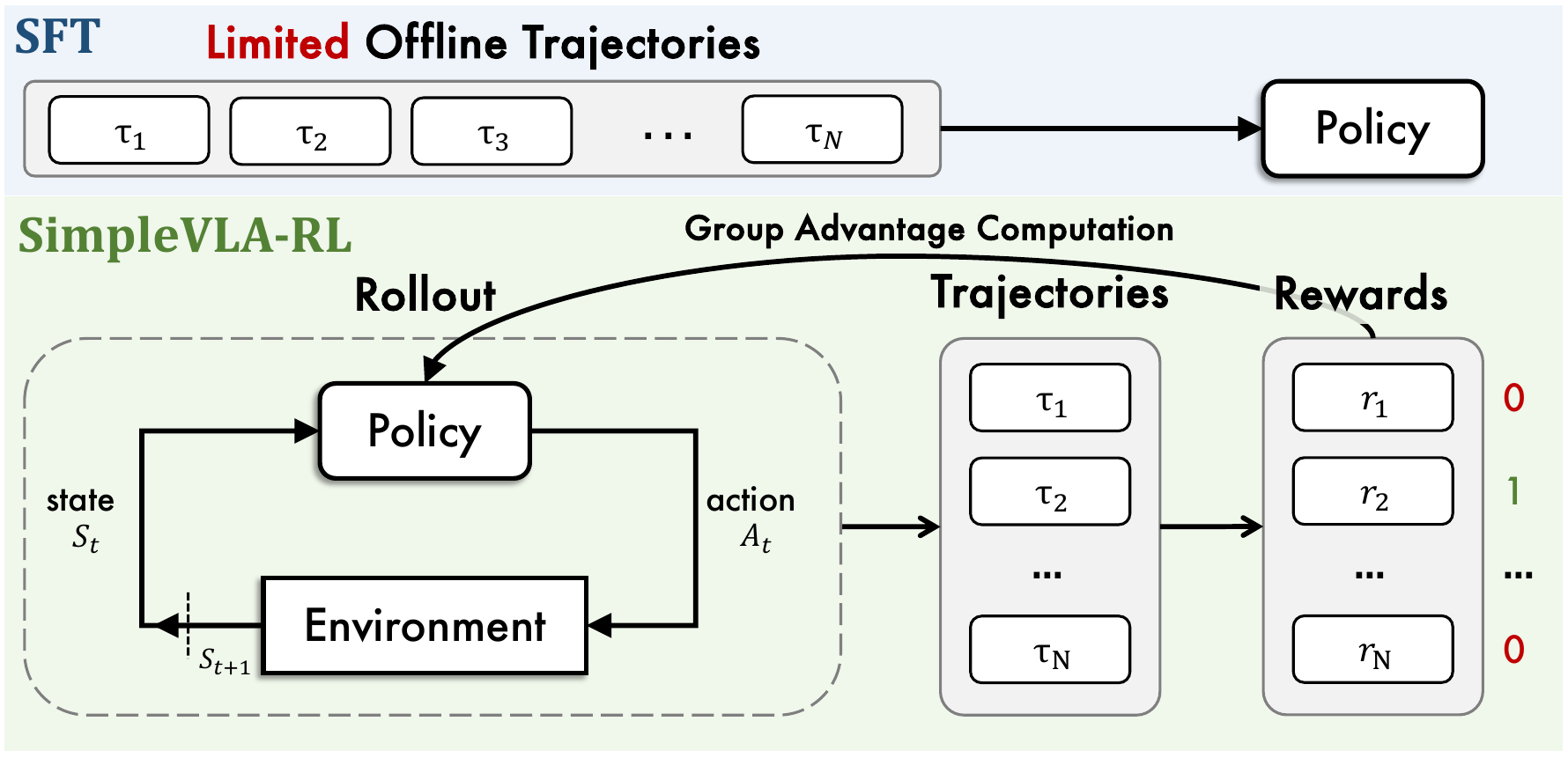}
  \caption{Overview of \method.}
  \label{fig:overview}
\end{figure}

DeepSeek-R1~\citep{guo2025deepseek} has achieved remarkable performance gains through online RL with the simple, scalable rule-based reward, highlighting a promising training paradigm.
In this section, we introduce \method, which extends this rule-based online RL framework to VLA models for embodied manipulation tasks as shown in Figure~\ref{fig:overview}.
Specifically, our training framework proceeds as follows: we begin by generating multiple trajectories for each input via random sampling~\textbf{($\S$\ref{sec:interactive_vla_rollout})}.
Each trajectory is then assigned a simple outcome reward (1 for success, 0 for failure) based on environment feedback~\textbf{($\S$~\ref{sec:outcome_reward})}.
Leveraging these rewards together with the corresponding action token probabilities, we compute the GRPO loss to update the policy model~\textbf{($\S$~\ref{sec:training_objective})}.

\subsection{Interactive VLA Rollout}
\label{sec:interactive_vla_rollout}

RL on VLA models differs fundamentally from LLMs in trajectory generation.
To enable online RL, policy models need to generate diverse trajectories from an input for effective exploration.
While LLMs naturally achieve this diversity through random sampling on text token distributions~\citep{renze2024effect,de2021survey}, VLA models face a unique challenge due to their action decoding strategies.
Current VLA models often employ three strategies:
(1) generating action token distributions similar to LLMs~\citep{black2024pi_0,kim2024openvla},
(2) diffusion-based denoising on latent states~\citep{liu2024rdt,cheang2025gr},
and (3) deterministic regression via MLPs~\citep{kim2025fine}.
Among these, the token-based approach is most compatible with PPO-like RL algorithms, as it naturally provides action distributions necessary for both random sampling and policy gradient computation.
Therefore, we adopt this approach, where the VLA model outputs action token probability distributions and employs random sampling to generate diverse trajectories.

Furthermore, for a given input query, LLM rollout proceeds by autoregressively generating tokens until reaching a stop token or max output length.
In contrast, VLA rollout requires continuous interaction with the environment to update the visual observation and robot state dynamically (as detailed in Section~\ref{sec:preliminaries}).
This closed-loop interaction is necessary because each robotic action alters the environment, and subsequent actions must be conditioned on real-time sensory feedback.
We present the comparison of the rollout algorithm pseudo-code of LLMs and VLA in Listing~\ref{code:rollout}.

\begin{figure}[!h]
\centering
\begin{tcolorbox}[
    colframe=gray,       %
    colback=white,       %
    boxrule=0.5pt,       %
    arc=2pt,             %
    left=0pt, right=4pt, top=1pt, bottom=2pt, %
    width=\linewidth, %
    enhanced
]
\noindent
\begin{normalcode}
def rollout(policy, dataset, number_sample=8, max_steps=None):
  rollout_dataset = []
  for batch in dataset:
     batch = batch.repeat(number_sample)
\end{normalcode}

\begin{diffdelcode}
-    # LLM generates diverse outputs using random sampling
-    outputs = policy.generate(batch, temperature=1.0)
-    rollout_dataset.append((batch, outputs))
\end{diffdelcode}

\begin{diffaddcode}
+    # Parallel env initialization and interaction
+    envs = env_process_pool.submit(batch.initialize)
+    states = env_process_pool.submit(envs.setup)
+    for t in range(max_steps):
+       # VLA generates diverse trajectories using temperature sampling on action tokens
+       actions = policy.generate(states, temperature=1.0)
+       rollout_dataset.append({f"{e.name}_step_{t}": (s,a) for e,s,a in zip(envs,states,actions)})
+       states, dones = env_process_pool.submit(envs.step, actions)
+       # Remove completed tasks
+       active = [(e,s) for e,s,d in zip(envs,states,dones) if not d]
+       if not active:
+          break
+       envs, states = zip(*active)
\end{diffaddcode}

\begin{normalcode}
  return rollout_dataset
\end{normalcode}
\end{tcolorbox}
\captionof{listing}{Pseudo-code for the adopted veRL rollout function: from LLM-based generation to interactive VLA sampling with synchronous environment parallelism.}
\label{code:rollout}
\end{figure}

\subsection{Outcome Reward Modeling}
\label{sec:outcome_reward}

\method employs a straightforward binary reward function for RL training.
Unlike traditional RL approaches that require carefully crafted reward functions~\citep{hadfield2017inverse,knox2023reward,booth2023perils}, we follow DeepSeek-R1's approach by assigning trajectory-level rewards of either 0 or 1 based solely on task completion.
When the VLA model successfully completes a task, the entire trajectory is assigned a reward of 1; otherwise, it receives a reward of 0.
For gradient computation, these trajectory-level rewards are uniformly propagated to the individual action tokens.
Consequently, all tokens within successful trajectories are assigned a reward of 1, whereas those in unsuccessful trajectories are assigned a reward of 0.
Our reward function is:
\begin{equation}
\label{eq:reward_func}
R(a_{i,t} \mid s_{i,t}) =
\begin{cases} 
1, & \text{is\_successful}[\text{traj}_i(a_i,s_i)], \\
0, & \text{otherwise}.
\end{cases}
\end{equation}
This simple outcome-level reward is simple yet effective: scalable, broadly applicable across environments, and free from complex process-based design~\citep{wu2021learning}.
By focusing solely on task completion, it avoids the non-transferability issues typical of task-specific rewards.

\subsection{Exploration Enhancements}
\label{sec:exploration}

Previous works~\citep{yu2025dapo,liu2025prorl,liu2025acereason,Polaris2025} have demonstrated that encouraging exploration during RL is critical.
We observe that this factor becomes even more crucial in VLA RL.
Manipulation tasks typically allow for a wide range of valid solutions.
However, VLA models tend to converge on a narrow set of solution patterns, largely due to the homogeneity of their training trajectories, which limits the efficiency of RL.
Promoting exploration encourages models to discover novel strategies and broaden the solution space, a property that is particularly advantageous in scenarios with low success rates.
Building on this insight, we implement three key modifications to enhance the exploration of RL training:
1) employing dynamic sampling during trajectory rollout,
2) adjusting the clip range in the GRPO training objective,
3) and increasing the sampling temperature during rollout.

\textbf{Dynamic Sampling\ \ }
Critic-free RL algorithms suffer from vanishing gradients when trajectories are assigned the same rewards.
For example, GRPO computes advantages using group-relative normalization, comparing each response’s reward to the mean and standard deviation of rewards within its group of sampled outputs.
When all trajectories share identical rewards, their advantage estimation becomes zero, resulting in null gradients and causing unstable training dynamics.

We address this challenge through Dynamic Sampling~\citep{yu2025dapo, cui2025process}, a method that has been proven effective in LLM RL~\citep{cui2025process,yu2025dapo,team2025kimi,shi2025efficient}.
During rollout, we exclude groups in which all trajectories either succeed or fail.
Sampling proceeds until the batch consists solely of groups with mixed outcomes, which can be formally expressed as:
\begin{equation}
0 < \left| \{\text{traj}_i(a_i, s_i) \mid \text{is\_successful}[\text{traj}_i(a_i, s_i)]\} \right| < G.
\end{equation}
This ensures non-zero advantage estimates and stable gradient flow throughout training.

\textbf{Clipping Higher\ \ } PPO and GRPO employ clipping over the importance sampling ratio to restrict the trust region~\citep{schulman2015trust} and enhance RL stability~\citep{schulman2017proximal,shao2024deepseekmath}.
However, the upper clipping threshold restricts the probability increase of low-probability tokens, thereby potentially constraining exploration.
Following DAPO~\citep{yu2025dapo}, we modify the clipping range in the GRPO training objective from [0.8, 1.2] to [0.8, 1.28].

\begin{figure}[htbp]
    \centering
    \begin{subfigure}[b]{0.32\textwidth}
        \centering
        \includegraphics[width=\textwidth]{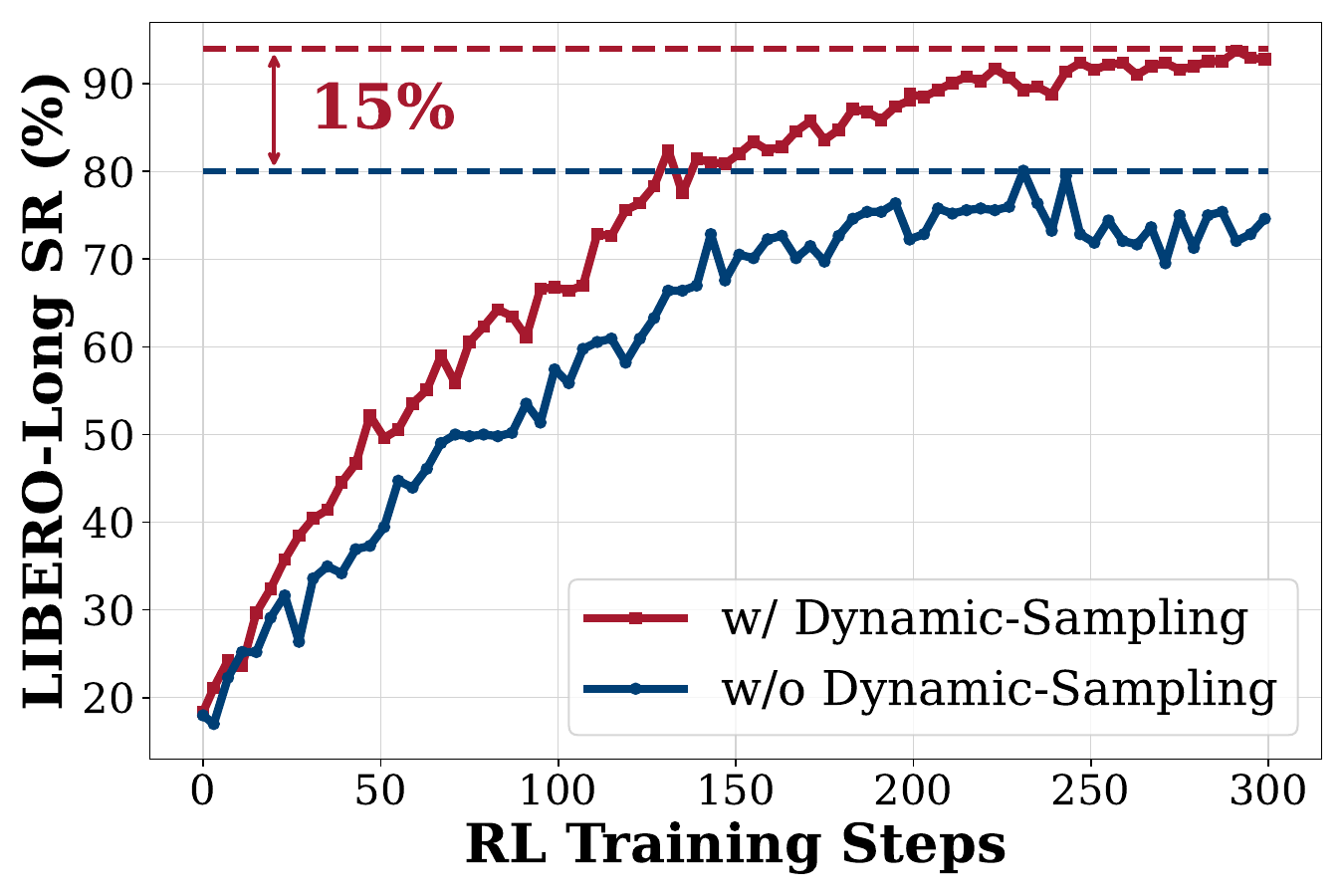}
        \caption{Dynamic Sampling}
        \label{fig:vla_ablation_3_dynamic_sampling}
    \end{subfigure}
    \hfill
    \begin{subfigure}[b]{0.32\textwidth}
        \centering
        \includegraphics[width=\textwidth]{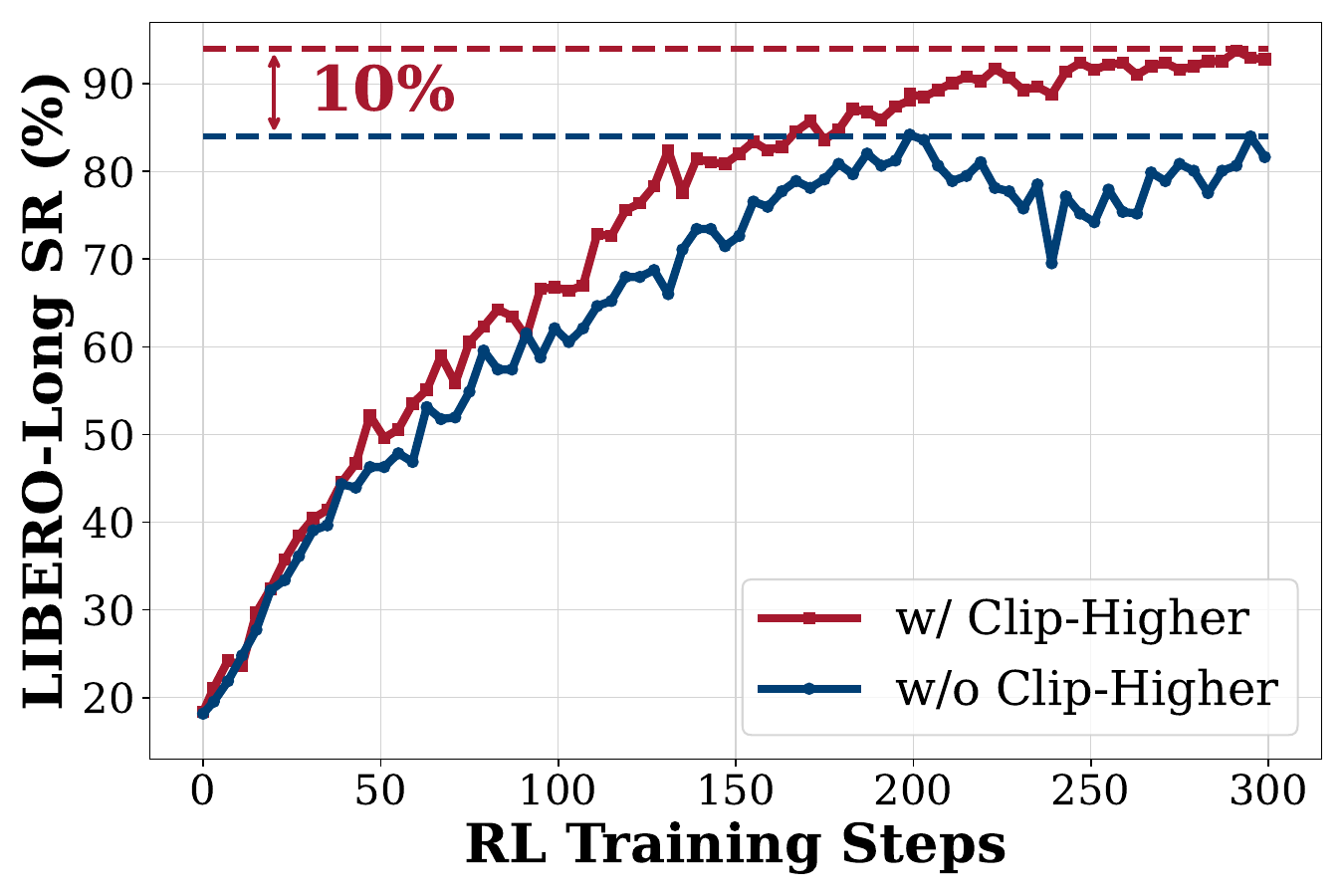}
        \caption{Clip Higher}
        \label{fig:vla_ablation_1_clip_higher}
    \end{subfigure}
    \hfill
    \begin{subfigure}[b]{0.32\textwidth}
        \centering
        \includegraphics[width=\textwidth]{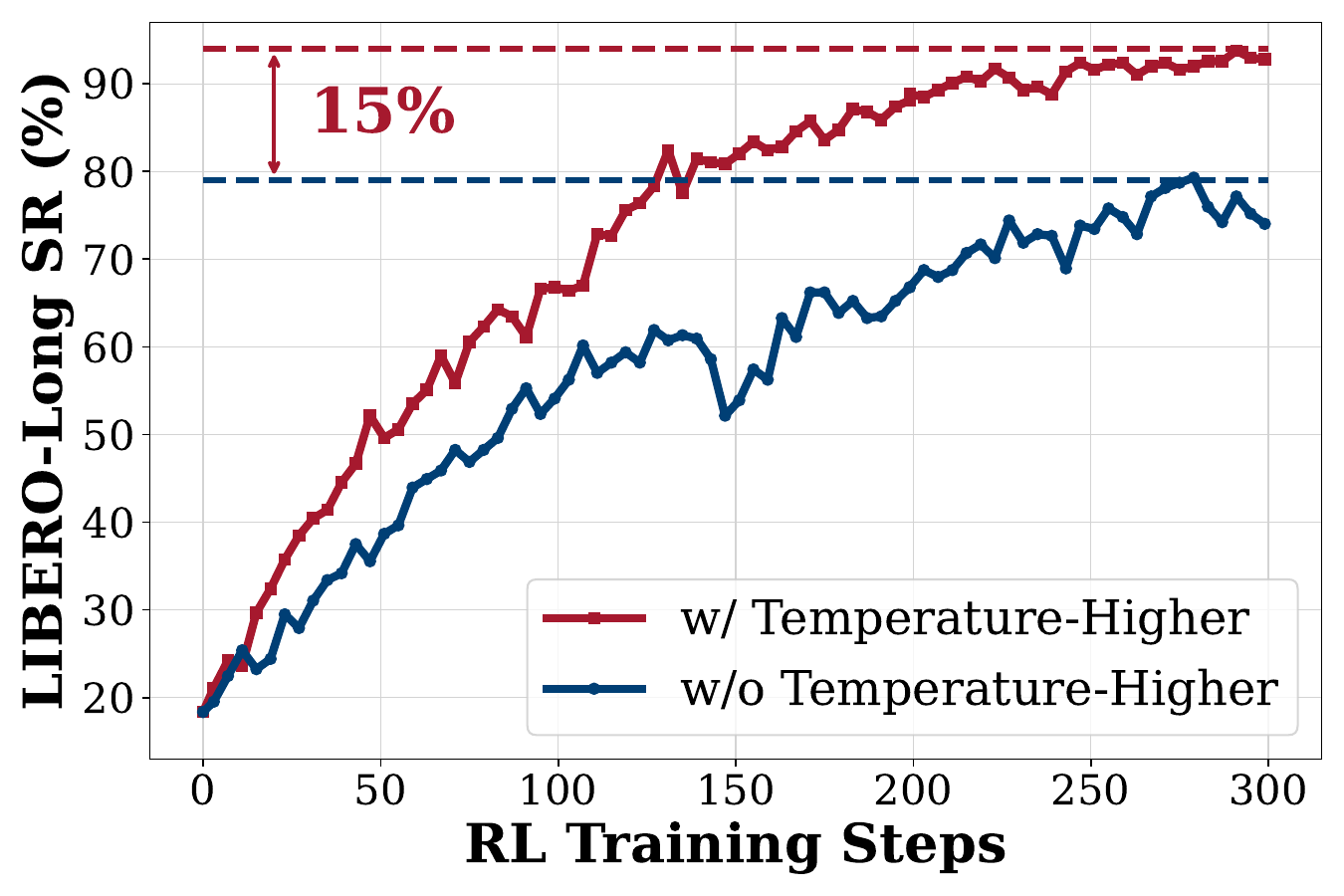}
        \caption{Higher Rollout Temperature}
        \label{fig:vla_ablation_2_temperature_higher}
    \end{subfigure}

    \caption{The effectiveness of three key enhancements: dynamic sampling, higher rollout temperature, and clip higher.
    }
    \label{fig:vla_ablation_combined}
\end{figure}

\textbf{Higher Rollout Temperature\ \ }
Recent works on LLM RL adjusting the rollout temperature to promote exploration have been widely shown to be effective, with sampling at higher temperatures yielding particularly notable improvements~\citep{Polaris2025,liu2025acereason,liao2025enhancing}.
To encourage the VLA model to generate more diverse trajectories during the rollout phase, we increase the sampling temperature from 1.0 to 1.6.
As shown in Figure \ref{fig:vla_ablation_combined}, these modifications led to notable improvements.

\subsection{Training Objective}
\label{sec:training_objective}

We use the adopted GRPO algorithm~\citep{shao2024deepseekmath} for online RL training on VLA models, with modifications as introduced in Section~\ref{sec:exploration}.
Moreover, we remove the KL divergence regularization following DAPO~\citep{yu2025dapo}.
This eliminates the need for a reference model during training, reducing memory consumption and accelerating the training.
Additionally, the KL penalty constrains policy divergence from a fixed reference, potentially limiting exploration of new behaviors.
Therefore, the policy is optimized via the following objective:
\begin{equation}
\label{eq:ours_loss}
\begin{split}
\mathcal{J}(\theta) &= \mathbb{E}_{s_0 \sim \mathcal{D}, \{a_t\}_{i=1}^G \sim \pi_{\theta_\text{old}}(\cdot|s_t)} \left[
\frac{1}{G} %
\sum_{i=1}^G \frac{1}{|a_i|} \sum_{t=1}^{|a_i|}
\min \left( r_{i,t}(\theta) \hat{A}_{i}, \,
\text{clip} \left( r_{i,t}(\theta), 1 - \varepsilon_{low}, 1 + \varepsilon_{high} \right) \hat{A}_{i} \right)
\right] \\
&\quad \text{s.t.} \quad
0 < \left| \{\text{traj}_i(a_i, s_i) \mid \text{is\_successful}[\text{traj}_i(a_i, s_i)]\} \right| < G,
\end{split}
\end{equation}
where
\begin{equation}
\label{eq:pos_and_adv}
r_{i,t}(\theta) = 
\frac{\pi_{\theta}(a_{i,t} \mid s_{i,t})}{\pi_{\theta_\text{old}}(a_{i,t} \mid s_{i,t})}, 
\quad
\hat{A}_{i} = 
\frac{R_{i} - \text{mean}(\{R_{i}\}_{i=1}^G)}{\text{std}(\{R_{i}\}_{i=1}^G)}.
\end{equation}

\section{Experiments}
\label{sec:experiments}

\begin{table}[!t]
  \centering
  \caption{RoboTwin 2.0 task classification based on planning horizon and required steps.}
  \resizebox{.6\linewidth}{!}{
  \begin{tabular}{l|c|c|c}
    \toprule
    \textbf{Task Name} & \textbf{Steps} & \textbf{Horizon} & \textbf{Horizon Group} \\
    \midrule
    \multicolumn{4}{c}{\cellcolor{green!20}\textbf{Short Horizon Tasks (112-130 steps)}} \\
    \midrule
    lift\_pot & 112 & Short & \multirow{4}{*}{\makecell{Average: 121 steps\\Count: 4 tasks}} \\
    beat\_block\_hammer & 113 & Short & \\
    pick\_dual\_bottles & 127 & Short & \\
    place\_phone\_stand & 130 & Short & \\
    \midrule
    \multicolumn{4}{c}{\cellcolor{yellow!20}\textbf{Medium Horizon Tasks (151-223 steps)}} \\
    \midrule
    move\_can\_pot & 151 & Medium & \multirow{4}{*}{\makecell{Average: 176 steps\\Count: 4 tasks}} \\
    place\_a2b\_left & 155 & Medium & \\
    place\_empty\_cup & 174 & Medium & \\
    handover\_mic & 223 & Medium & \\
    \midrule
    \multicolumn{4}{c}{\cellcolor{orange!20}\textbf{Long Horizon Tasks (283-313 steps)}} \\
    \midrule
    handover\_block & 283 & Long & \multirow{2}{*}{\makecell{Average: 298 steps\\Count: 2 tasks}} \\
    stack\_bowls\_two & 313 & Long & \\
    \midrule
    \multicolumn{4}{c}{\cellcolor{red!20}\textbf{Extra Long Horizon Tasks (466-637 steps)}} \\
    \midrule
    blocks\_rank\_rgb & 466 & Extra-Long & \multirow{2}{*}{\makecell{Average: 552 steps\\Count: 2 tasks}} \\
    put\_bottles\_dustbin & 637 & Extra-Long & \\
    \midrule
    \multicolumn{1}{l|}{\textbf{Overall Statistics}} & \multicolumn{3}{c}{\textbf{Total: 12 tasks, Average: 256 steps}} \\
    \bottomrule
  \end{tabular}
  }
  \label{tab:task_horizon_analysis}
\end{table}

\subsection{Experimental Setup}

\textbf{Benchmarks\ \ }
We evaluate \method on the widely used simulation benchmark, LIBERO~\citep{liu2023libero}, RoboTwin1.0~\citep{mu2025robotwin}, and RoboTwin2.0~\citep{chen2025robotwin}, and conduct real-world experiments on RoboTwin2.0 tasks.
LIBERO is a lifelong learning benchmark focused on language-guided manipulation tasks across diverse object types, task specifications, and environments.
It consists of five task suites: LIBERO-Goal, LIBERO-Spatial, LIBERO-Object, LIBERO-Long (10 tasks, each with 50 expert demonstrations), and LIBERO-90 (90 tasks for large-scale multitask evaluation).
We evaluate performance using the average Success Rate~(SR) across 50 held-out test scenarios for each task.
Moreover, RoboTwin1.0~\citep{mu2025robotwin} and RoboTwin2.0~\citep{chen2025robotwin} are simulation benchmarks for dual-arm manipulation.
RoboTwin1.0 provides 17 bimanual tasks with limited scene and object diversity. RoboTwin2.0 extends to 50 tasks across multiple robot embodiments with 731 object instances and comprehensive domain randomization (clutter, lighting, background, tabletop height, and language instructions), which enhances task diversity and sim-to-real transfer.
For RoboTwin2.0 training and evaluation, we employ the Agilex Piper robotic arm and domain-randomized task settings, with each task evaluated on 100 held-out test scenarios.
For RoboTwin2.0, we select and categorize 12 tasks into 4 horizon levels based on the average number of steps per task for comprehensive evaluation.
Table~\ref{tab:task_horizon_analysis} summarizes the specific step counts for each task and the step ranges for different horizon levels of the various tasks in RoboTwin2.0.

\begin{table}[!t]
  \centering
  \caption{Main results of different VLA models on LIBERO.}
  \resizebox{.5\linewidth}{!}{
    \begin{tabular}{lccccc}
      \toprule
      \multirow{2}{*}{\textbf{Model}} &
        \multicolumn{5}{c}{\textbf{LIBERO}} \\
      \cmidrule(lr){2-6}
      & \textbf{Spatial} & \textbf{Object} & \textbf{Goal} & \textbf{Long} & \textbf{Avg} \\
      \midrule
      Octo                & 78.9 & 85.7 & 84.6 & 51.1 & 75.1 \\
      OpenVLA             & 84.7 & 88.4 & 79.2 & 53.7 & 76.5 \\
      Nora                & 92.2 & 95.4 & 89.4 & 74.6 & 87.9 \\
      $\pi_0$ + FAST      & 96.4 & 96.8 & 88.6 & 60.2 & 85.5 \\
      $\pi_0$             & 96.8 & 98.8 & 95.8 & 85.2 & 94.2 \\
      UniVLA              & 96.5 & 96.8 & 95.6 & 92.0 & 95.2 \\
      \midrule
      OpenVLA-OFT         & 91.6 & 95.3 & 90.6 & 86.5 & 91.0 \\
      \quad \textbf{w/ ours}       & \textbf{99.4} & \textbf{99.1} & \textbf{99.2} & \textbf{98.5} & \textbf{99.1} \\
      \rowcolor{lightblue!100} \quad $\Delta$ &
        \textcolor{red}{$+7.8$} &
        \textcolor{red}{$+3.8$} &
        \textcolor{red}{$+8.6$} &
        \textcolor{red}{$+12.0$} &
        \textcolor{red}{$+8.1$} \\
      \bottomrule
    \end{tabular}
  }
  \label{tab:main_baselines}
\end{table}

\begin{table}[!t]
  \centering
  \caption{Main results of different VLA models on RoboTwin1.0.}
  \resizebox{.8\linewidth}{!}{
    \begin{tabular}{lccccc}
      \toprule
      \multirow{2}{*}{\textbf{Model}} &
        \multicolumn{4}{c}{\textbf{RoboTwin1.0}} &
        \multirow{2}{*}{\textbf{Avg}} \\
      \cmidrule(lr){2-5}
      & \textbf{Hammer Beat} & \textbf{Block Handover} & \textbf{Blocks Stack} & \textbf{Shoe Place} & \\
      \midrule
      DP                  & 0.0  & 12.0 & 7.1    & 4.3  & 5.9 \\
      DP3                 & 64.7 & 84.3 & 24.0    & 59.3 & 58.1 \\
      \midrule
      OpenVLA-OFT & 67.2 & 61.6 & 7.1  & 23.4 & 39.8 \\
      \quad \textbf{w/ ours} & \textbf{92.6} & \textbf{89.6} & \textbf{40.2} & \textbf{59.3} & \textbf{70.4} \\
      \rowcolor{lightblue!100} \quad $\Delta$ &
        \textcolor{red}{$+25.4$} &
        \textcolor{red}{$+28.0$} &
        \textcolor{red}{$+33.1$} &
        \textcolor{red}{$+35.9$} &
        \textcolor{red}{$+30.6$} \\
      \bottomrule
    \end{tabular}
  }
  \label{tab:robotwin1_results}
\end{table}

\textbf{Backbones\ \ }
We apply \method to OpenVLA-OFT~\citep{kim2025fine}, a state-of-the-art auto-regressive VLA model that achieves high performance and inference efficiency.
Based on OpenVLA~\citep{kim2024openvla}, it employs vision encoders and LLaMA2-7B~\citep{touvron2023llama} as the backbone with action chunk and parallel decoding designs, making it particularly suitable for online RL where model inference is frequent.
Our implementation of the OpenVLA-OFT model differs from the official version.
To achieve improved training and inference efficiency, we utilize only single-view images, language instructions, and robot proprioceptive states as model inputs, whereas the official model additionally incorporates wrist camera images.
Additionally, in the LIBERO, we don't use robot proprioceptive states in model inputs.
Regarding the model architecture, we employ only parallel decoding and action chunking designs.
We use the LLaMA2 output head to generate action tokens and the cross-entropy loss, whereas the official model uses an MLP to generate continuous actions and L1 regression.
Due to the differences in model inputs and architecture, we cannot use the official checkpoints. We modify the official codebase and performed SFT from scratch using the same datasets and hyperparameters as the official implementation.

\textbf{Baselines\ \ } We compare with recent advanced VLA models: UniVLA~\citep{bu2025univla}, RDT-1B~\citep{liu2024rdt}, $\pi_0$~\citep{black2024pi_0}, $\pi_\text{fast}$~\citep{pertsch2025fast}, Nora~\citep{hung2025nora}, OpenVLA~\citep{kim2024openvla}, Octo~\citep{team2024octo}, DP~\citep{chi2024diffusionpolicy} and DP3~\citep{ze20243d}.

\textbf{Implementation Details\ \ }
For training infrastructure, we employ 8 x NVIDIA A800 80GB for full-parameter training.
The training hyperparameters are configured as follows: learning rate $lr = 5 \times 10^{-6}$, training batch size of 64, sampling count of 8, mini-batch size of 128, clip ratio $\varepsilon_{low} = 0.2$, $\varepsilon_{high} = 0.28$, and temperature $T = 1.6$. The number of action chunks is 8 in the LIBERO and 25 in the RoboTwin1.0\&2.0.
The model is configured with a total of 256 action tokens.
The maximum environment interaction step is set to 512 in the LIBERO and 200, 400, or 800 in the RoboTwin1.0\&2.0, depending on different tasks.
During the rollout phase of RL training, we employ random sampling.
For evaluation, we utilize greedy sampling, with each benchmark tested three times for reproducibility.

\subsection{Main Results}

\begin{table}[!t]
  \centering
  \caption{Main results of different VLA models on RoboTwin2.0, organized by task horizon.}
  \resizebox{\linewidth}{!}{
    \begin{tabular}{lccccc}
      \toprule
      \multicolumn{6}{c}{\textbf{Short Horizon Tasks (100-130 Steps)}} \\
      \midrule
      \textbf{Model} & \textbf{Lift Pot} & \textbf{Beat Hammer Block} & \textbf{Pick Dual Bottles} & \textbf{Place Phone Stand} & \textbf{Avg} \\
      \midrule
      $\pi_0$ & 51.0 & 59.0 & 50.0 & 22.0 & 45.5 \\
      RDT & 45.0 & 22.0 & 18.0 & 13.0 &  24.5 \\
      \midrule
      OpenVLA-OFT & 10.1 & 28.1 & 29.7 & 17.1 & 21.3 \\
      \quad \textbf{w/ ours} & \textbf{64.1} & \textbf{87.5} & \textbf{68.3} & \textbf{39.6} & \textbf{64.9} \\
      \rowcolor{lightblue!100} \quad $\Delta$ & \textcolor{red}{$+54.0$} & \textcolor{red}{$+59.4$} & \textcolor{red}{$+38.6$} & \textcolor{red}{$+22.5$} & \textcolor{red}{$+43.6$} \\
      \midrule
      \multicolumn{6}{c}{\textbf{Medium Horizon Tasks (150-230 Steps)}} \\
      \midrule
      \textbf{Model} & \textbf{Move Can Pot} & \textbf{Place A2B Left} & \textbf{Place Empty Cup} & \textbf{Handover Mic} & \textbf{Avg} \\
      \midrule
      $\pi_0$ & 41.0 & 38.0 & 60.0 & 96.0 & 58.8 \\
      RDT & 33.0 & 21.0 & 42.0 & 95.0 &  47.8 \\
      \midrule
      OpenVLA-OFT & 28.1 & 37.5 & 77.3 & 45.3 & 47.1 \\
      \quad \textbf{w/ ours} & \textbf{61.2} & \textbf{45.3} & \textbf{94.2} & \textbf{89.2} & \textbf{72.5} \\
      \rowcolor{lightblue!100} \quad $\Delta$ & \textcolor{red}{$+33.1$} & \textcolor{red}{$+7.8$} & \textcolor{red}{$+16.9$} & \textcolor{red}{$+43.9$} & \textcolor{red}{$+25.4$} \\
      \midrule
      \multicolumn{6}{c}{\textbf{Long (280-320 Steps) \& Extra Long Horizon Tasks (450-650 Steps)}} \\
      \midrule
      \textbf{Model} & \textbf{Handover Block} & \textbf{Stack Bowls Two} & \textbf{Blocks Rank Rgb} & \textbf{Put Bottles Dustbin} & \textbf{Avg} \\
      \midrule
      $\pi_0$ & 39.0 & 53.0 & 45.0 & 36.0 & 43.3 \\
      RDT & 26.0 & 42.0 & 17.0 & 26.0 &  27.8 \\
      \midrule
      OpenVLA-OFT & 33.1 & 40.6 & 70.2 & 42.2 & 46.5 \\
      \quad \textbf{w/ ours} & \textbf{57.8} & \textbf{75.8} & \textbf{81.3} & \textbf{60.9} & \textbf{69.0} \\
      \rowcolor{lightblue!100} \quad $\Delta$ & \textcolor{red}{$+24.7$} & \textcolor{red}{$+35.2$} & \textcolor{red}{$+11.1$} & \textcolor{red}{$+18.7$} & \textcolor{red}{$+22.4$} \\
      \midrule
      \textbf{Overall Avg} & \multicolumn{4}{r}{\textbf{RDT:} 33.3  \quad \quad $\mathbf{\pi_0}$: 49.2 \quad \quad  \textbf{OpenVLA-OFT:} 38.3 \quad \textbf{w/ ours}: 68.8} & \textcolor{red}{$+30.5$} \\
      \bottomrule
    \end{tabular}
  }
  \label{tab:robotwin2_results_by_difficulty}
\end{table}

We evaluate~\method on the three benchmarks: LIBERO, RoboTwin1.0, and RoboTwin2.0.
For each benchmark, we employ a two-stage training paradigm in which SFT is followed by \method, both applied to OpenVLA-OFT.
In contrast, the baseline models are trained solely with SFT.
For the four task suites of LIBERO, we perform SFT using all 500 demonstrations per task suite, then perform RL on 500 corresponding simulation scenarios.
For RoboTwin1.0, we use 50 demonstrations per task for single-task SFT, then perform RL on 100 simulation scenarios per task.
For RoboTwin2.0, we use 1,000 demonstrations per task for SFT, then perform RL on 1,000 scenarios per task.

Tables~\ref{tab:main_baselines}, \ref{tab:robotwin1_results}, and \ref{tab:robotwin2_results_by_difficulty} present the results on the three benchmarks: LIBERO, RoboTwin1.0, and RoboTwin2.0, respectively.
On four LIBERO task suites, \method improves the average success rate of the SFT-tuned OpenVLA-OFT model from 91\% to 99\%, achieving SoTA performance and surpassing all current  VLA models such as $\pi_0$ and UniVLA.
For long-horizon tasks in LIBERO-Long, \method achieves 98.5\% success rate, with a 12\% improvement over the baseline (86.5\%) and 13.3\% gains over $\pi_0$ (85.2\%).
The results demonstrate that \method can substantially improve model performance even when SFT has already achieved strong results.
On RoboTwin1.0's four dual-arm tasks, \method achieves a 30.6\% gain over the fine-tuned OpenVLA-OFT baseline (from 39.8\% to 70.4\%).
Across 12 dual-arm tasks of RoboTwin2.0, \method achieves an 80\% relative improvement, lifting performance from 38.3\% to 68.8\% and outperforming SoTA methods including $\pi_0$ (49.2\%) and RDT (33.3\%).
Notably, even on two Extra-Long-Horizon tasks requiring multi-round dual-arm interactions such as ``Blocks Rank Rgb'' and ``Put Bottles Dustbin'', \method achieves 11.1\% and 18.7\% points gain, respectively.
\method shows consistent improvements in all horizon levels, validating the effectiveness of outcome-level reward even for complex long-horizon tasks.
The results demonstrate that \method consistently improves model performance across diverse benchmarks without requiring additional demonstration data.

\section{Analysis}

In this section, we analyze the role of \method in addressing three key challenges that hinder the further advancement and scaling of the VLA model: \textbf{data, generalization, and real-world tasks}.
Below are several key takeaways:

\begin{tcolorbox}[takeawaysbox]
\begin{enumerate}[leftmargin=1em]
    \item \textbf{Data:} \method can significantly reduce reliance on demonstration data, effectively alleviating the data scarcity bottleneck that constrains VLA scaling~\textbf{($\S$~\ref{sec:ana_data})}.
    \item \textbf{Generalization:} Compared to SFT, \method demonstrates strong generalization in spatial configurations, object types, and task settings~\textbf{($\S$~\ref{sec:generalization})}.
    \item \textbf{Real-world Task:} \method exhibits strong sim-to-real transfer, with large-scale simulation training remarkably improving real-world performance, indicating a promising path for scaling up real-world policy~\textbf{($\S$~\ref{sec:real_world})}.
\end{enumerate}
\end{tcolorbox}

\subsection{Overcoming Data Scarcity}
\label{sec:ana_data}

Developing foundation VLA models for manipulation tasks requires large-scale demonstration data for training~\citep{liu2024rdt,black2024pi_0,intelligence2025pi_}.
This data scaling paradigm has been proven in the NLP area~\citep{hoffmann2022training,achiam2023gpt,touvron2023llama}.
However, acquiring high-quality trajectory data for embodied manipulation tasks remains expensive and difficult, creating a fundamental bottleneck for VLA model development~\citep{bi2025h,zhong2025survey}.
Therefore, we investigate whether~\method can enhance VLA models even with extremely limited demonstration trajectories to overcome this limitation.

\textbf{Settings\ } To simulate scenarios with scarce demonstration data, we finetune OpenVLA-OFT using only one demonstration data per task, denoted as \textit{One-Trajectory SFT}.
Given that each of the four LIBERO task suites contains 10 distinct tasks, we utilize merely 10 demonstration data per task suite.
For comparison, we also conduct an experiment using all available demonstration data for each task, 500 per task suite, denoted as \textit{Full-Trajectory SFT}.
Following both \textit{One-Trajectory SFT} and \textit{Full-Trajectory SFT}, we apply \method on the SFT model.

\begin{wraptable}{r}{0.57\textwidth}
  \centering
  \caption{Comparisons between One-Trajectory and Full-Trajectory SFT on LIBERO.}
  \resizebox{\linewidth}{!}{
    \begin{tabular}{lccccc}
      \toprule
      \multirow{2}{*}{\textbf{Model}} &
        \multicolumn{5}{c}{\textbf{LIBERO}} \\
      \cmidrule(lr){2-6}
      & \textbf{Spatial} & \textbf{Object} & \textbf{Goal} & \textbf{Long} & \textbf{Avg} \\
      \midrule
      \multicolumn{6}{c}{One-Trajectory SFT} \\
      \midrule
      OpenVLA-OFT         & 63.6 & 54.9 & 59.6 & 17.3 & 48.9 \\
      \quad \textbf{w/ ours}       & \textbf{98.2} & \textbf{98.7} & \textbf{98.8} & \textbf{91.7} & \textbf{96.9} \\
      \rowcolor{lightblue!100} \quad $\Delta$ &
        \textcolor{red}{$+34.6$} &
        \textcolor{red}{$+43.8$} &
        \textcolor{red}{$+39.2$} &
        \textcolor{red}{$+74.4$} &
        \textcolor{red}{$+48.0$} \\
      \midrule
      \multicolumn{6}{c}{Full-Trajectory SFT} \\
      \midrule
      OpenVLA-OFT         & 91.6 & 95.3 & 90.6 & 86.5 & 91.0 \\
      \quad \textbf{w/ ours}       & \textbf{99.4} & \textbf{99.1} & \textbf{99.2} & \textbf{98.5} & \textbf{99.1} \\
      \rowcolor{lightblue!100} \quad $\Delta$ &
        \textcolor{red}{$+7.8$} &
        \textcolor{red}{$+3.8$} &
        \textcolor{red}{$+8.6$} &
        \textcolor{red}{$+12.0$} &
        \textcolor{red}{$+8.1$} \\
      \bottomrule
    \end{tabular}
  }
  \label{tab:data_scare}
\end{wraptable}

\textbf{Results\ } As shown in Table~\ref{tab:data_scare}, the limitations of SFT become notable when demonstration data is limited.
Compared to results under \textit{Full-Trajectory SFT} setting, the SFT success rates on LIBERO-Spatial, LIBERO-Object, and LIBERO-Goal fall below 63.6\%, while the performance on long-horizon tasks in LIBERO-Long drops to only 17.3\%, highlighting the challenge of relying solely on SFT for long-horizon tasks.
Remarkably, after applying \method on \textit{One-Trajectory SFT}, the average success rate across the four LIBERO task suites increases from 48.9\% to 96.9\%, even surpassing the 91\% of \textit{Full-Trajectory SFT}.
Notably, LIBERO-Long improves from 17.3\% to 91.7\%, while the other three task suites all exceed 98\% success rate, each exceeding their \textit{Full-Trajectory SFT} performance.
The performance gap between \textit{One-Trajectory SFT + RL} (96.9\%) and \textit{Full-Trajectory SFT + RL} (99.1\%) is only 2.2\%.
These results demonstrate that \method can substantially improve VLA model performance in data-scarce scenarios. 
This suggests that through trial-and-error exploration with outcome feedback, online RL methods such as \method can enable further scaling of VLA model training, even when only minimal demonstration data is available.

\subsection{Generalization Analysis}\label{sec:generalization}

The generalization ability of VLA models remains a key challenge~\citep{intelligence2025pi_,zhong2025survey,liu2025can}.
This subsection evaluates how SFT and online RL methods like \method affect VLA generalization across three dimensions: spatial (LIBERO-Spatial), objects (LIBERO-Object), and tasks (LIBERO-Goal).

\textbf{Settings\ }
We conduct this experiment using three task suites, LIBERO-Spatial, LIBERO-Object, and LIBERO-Goal, each of which comprises ten distinct tasks.
For each LIBERO task suite, we randomly select nine tasks as seen tasks for RL or SFT training, while reserving the remaining task as the unseen task for out-of-distribution evaluation.
For both methods, we first fine-tune OpenVLA-OFT under the \textit{One-Trajectory SFT} setting to obtain a base model with non-zero success rates, since the original model achieves only 0\% on LIBERO and is incapable of performing online RL.
For SFT, we further fine-tune the \textit{One-Trajectory SFT} base model (\S \ref{sec:ana_data}) using all 450 demonstration trajectories from 9 seen tasks on each task suite.
Moreover, we perform \method on this \textit{One-Trajectory SFT} base model.
We plot how the models' performance on unseen tasks changes as their training tasks' success rates increase during training.

\begin{figure}[!tbp]
    \centering
    \vspace{-0.2cm}

    \begin{subfigure}[b]{0.32\textwidth}
        \centering
        \includegraphics[width=\textwidth]{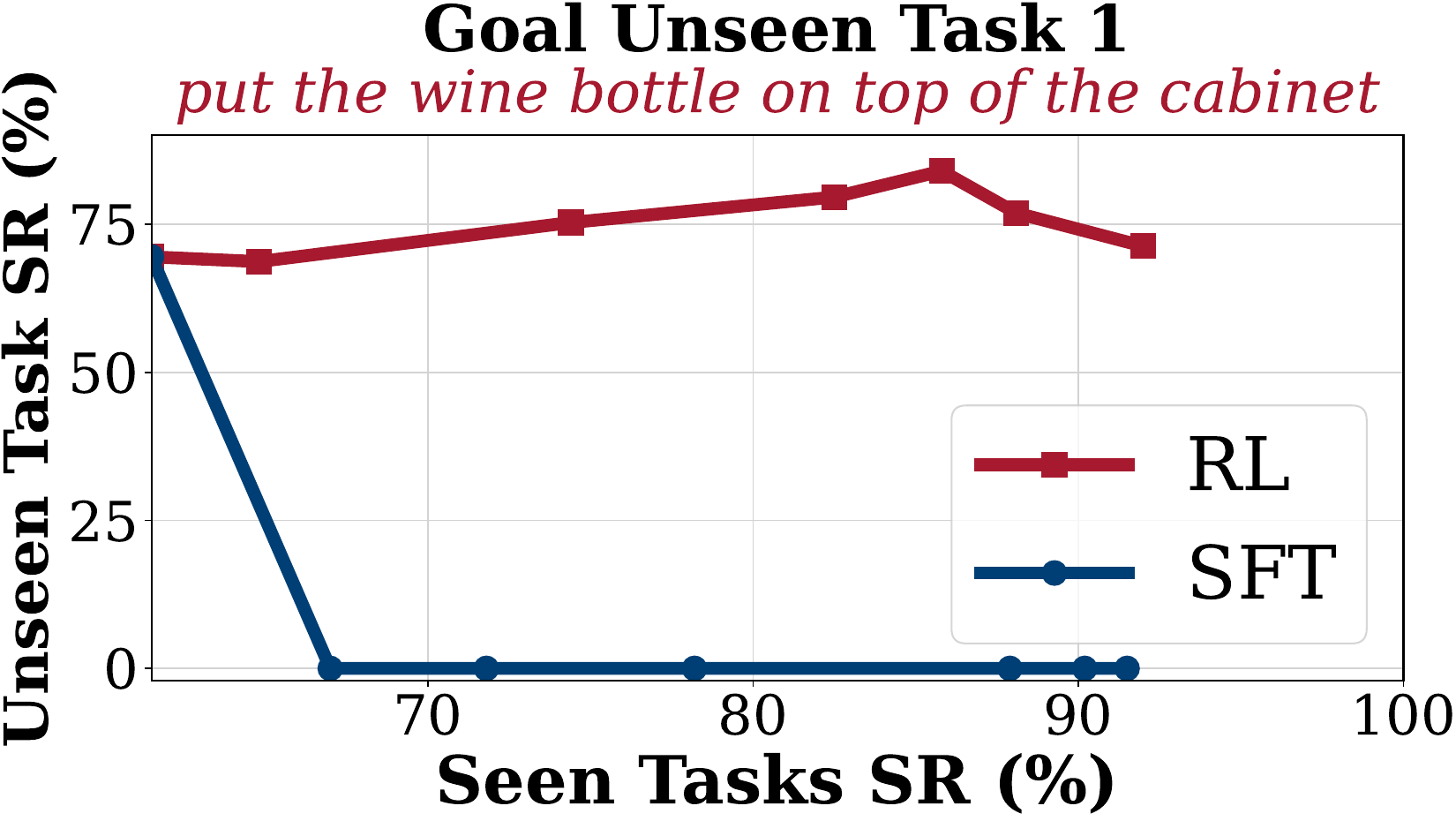}
        \label{fig:libero_goal_a}
    \end{subfigure}
    \hfill
    \begin{subfigure}[b]{0.32\textwidth}
        \centering
        \includegraphics[width=\textwidth]{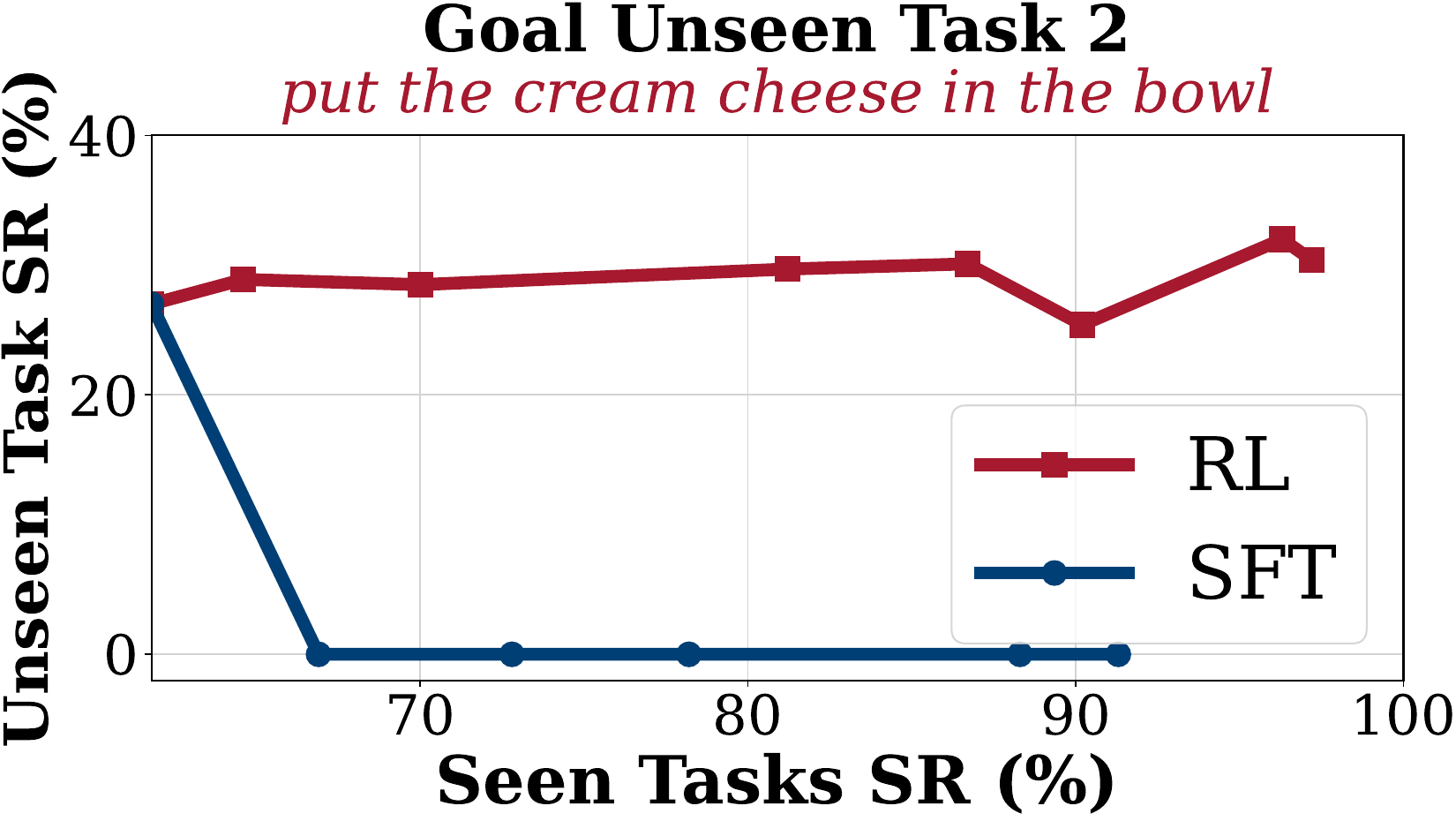}
        \label{fig:libero_goal_b}
    \end{subfigure}
    \hfill
    \begin{subfigure}[b]{0.32\textwidth}
        \centering
        \includegraphics[width=\textwidth]{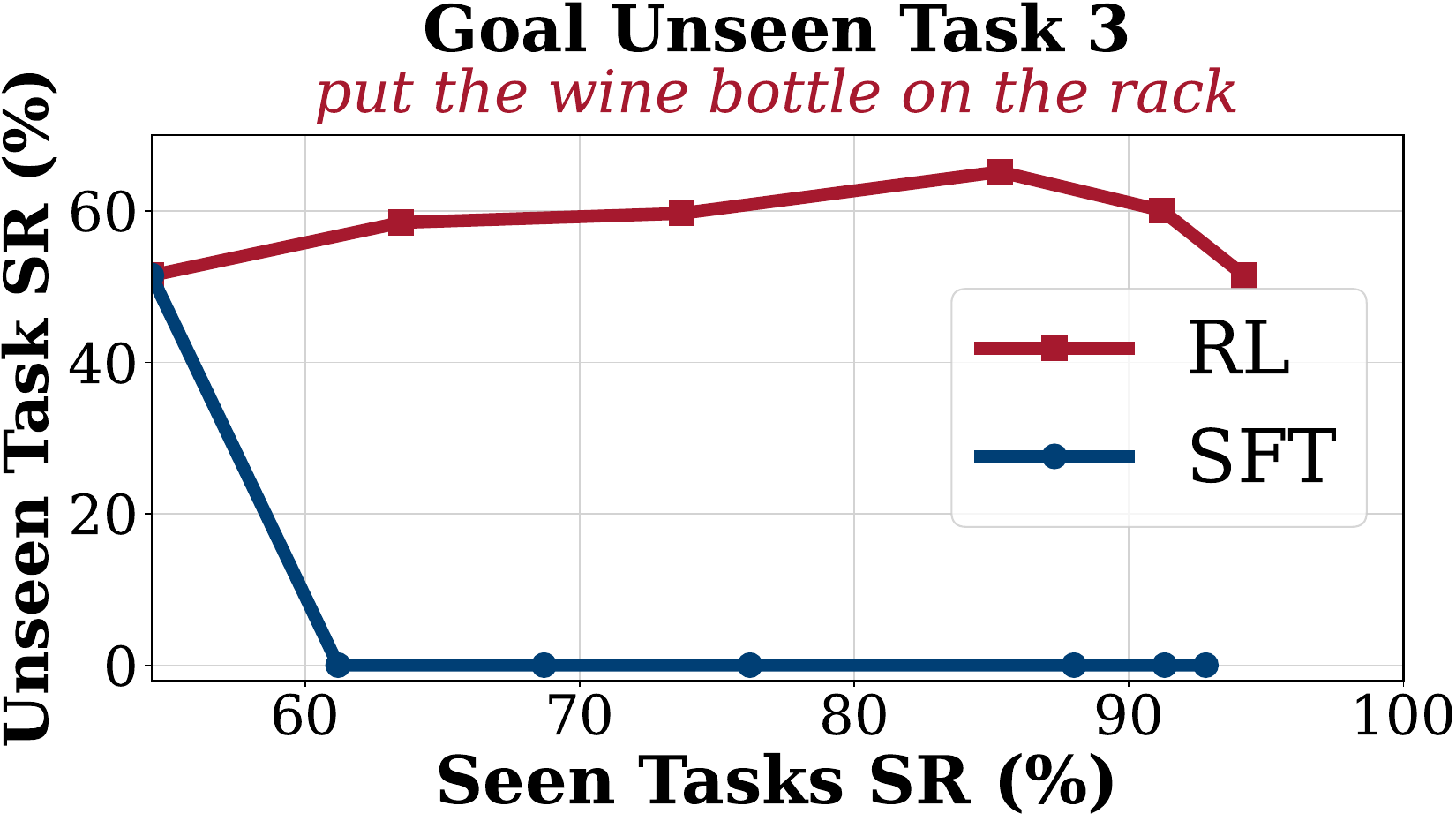}
        \label{fig:libero_goal_c}
    \end{subfigure}

    \vspace{-0.2cm}

    \begin{subfigure}[b]{0.32\textwidth}
        \centering
        \includegraphics[width=\textwidth]{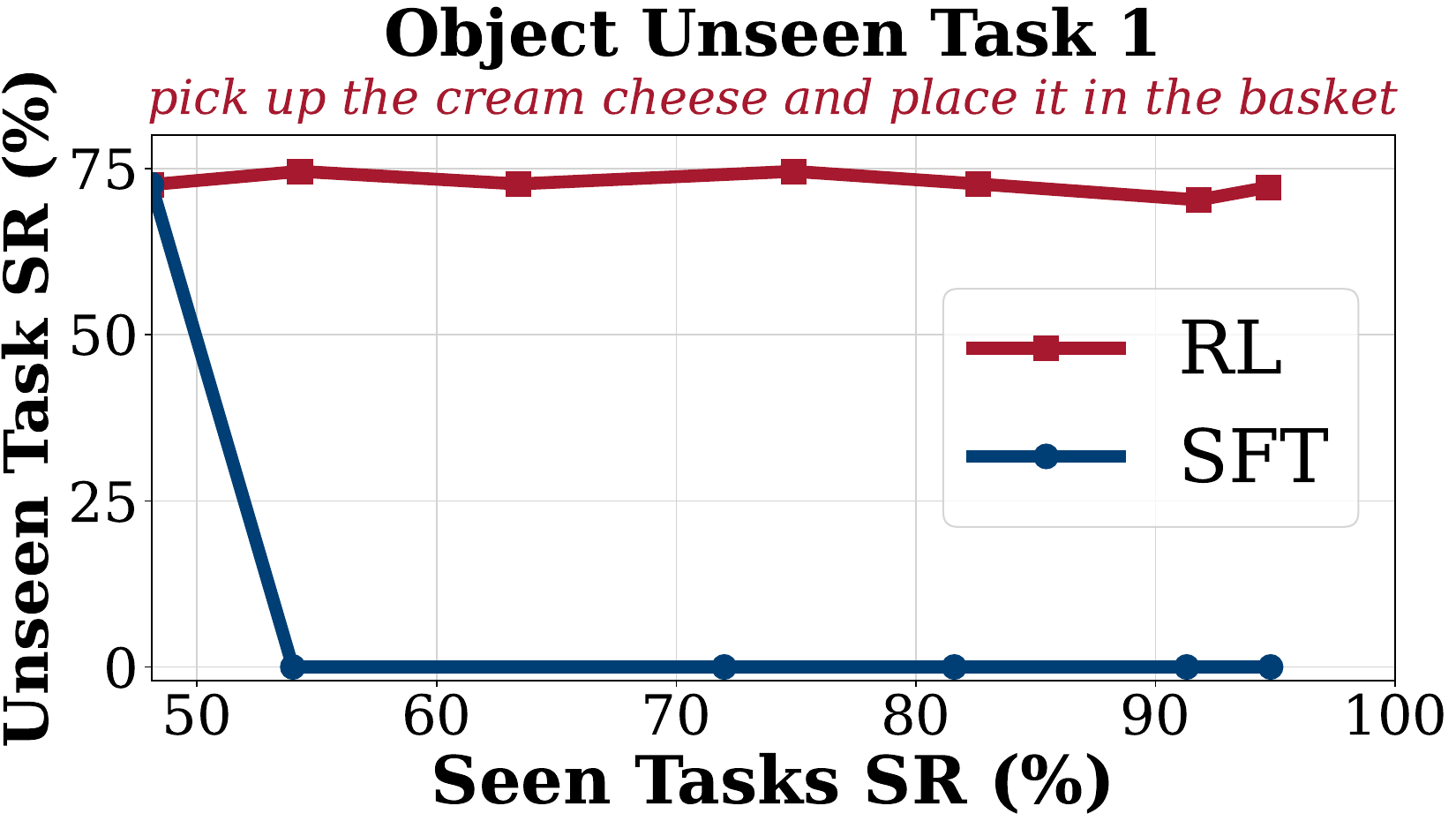}
        \label{fig:libero_object_a}
    \end{subfigure}
    \hfill
    \begin{subfigure}[b]{0.32\textwidth}
        \centering
        \includegraphics[width=\textwidth]{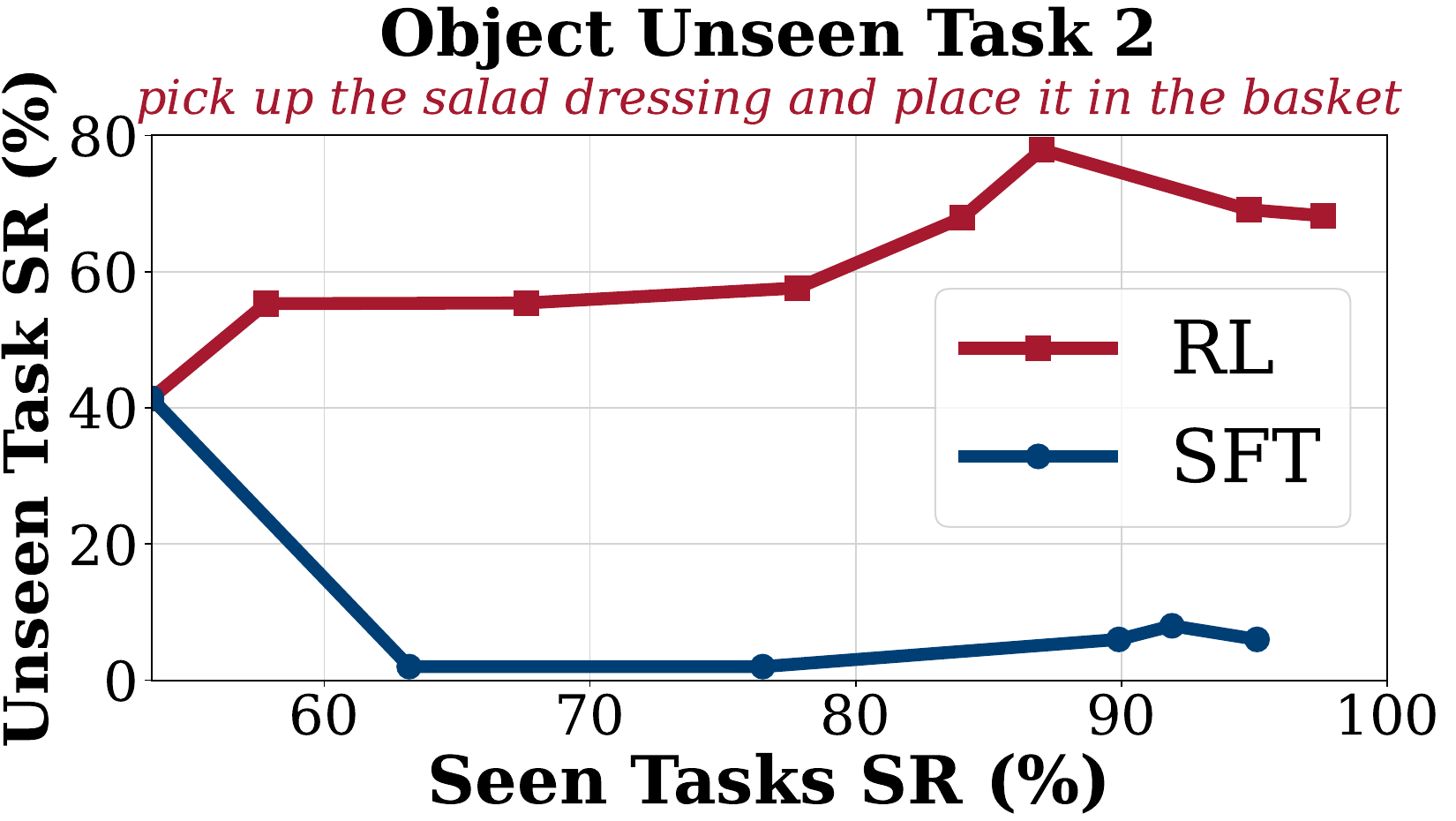}
        \label{fig:libero_object_b}
    \end{subfigure}
    \hfill
    \begin{subfigure}[b]{0.32\textwidth}
        \centering
        \includegraphics[width=\textwidth]{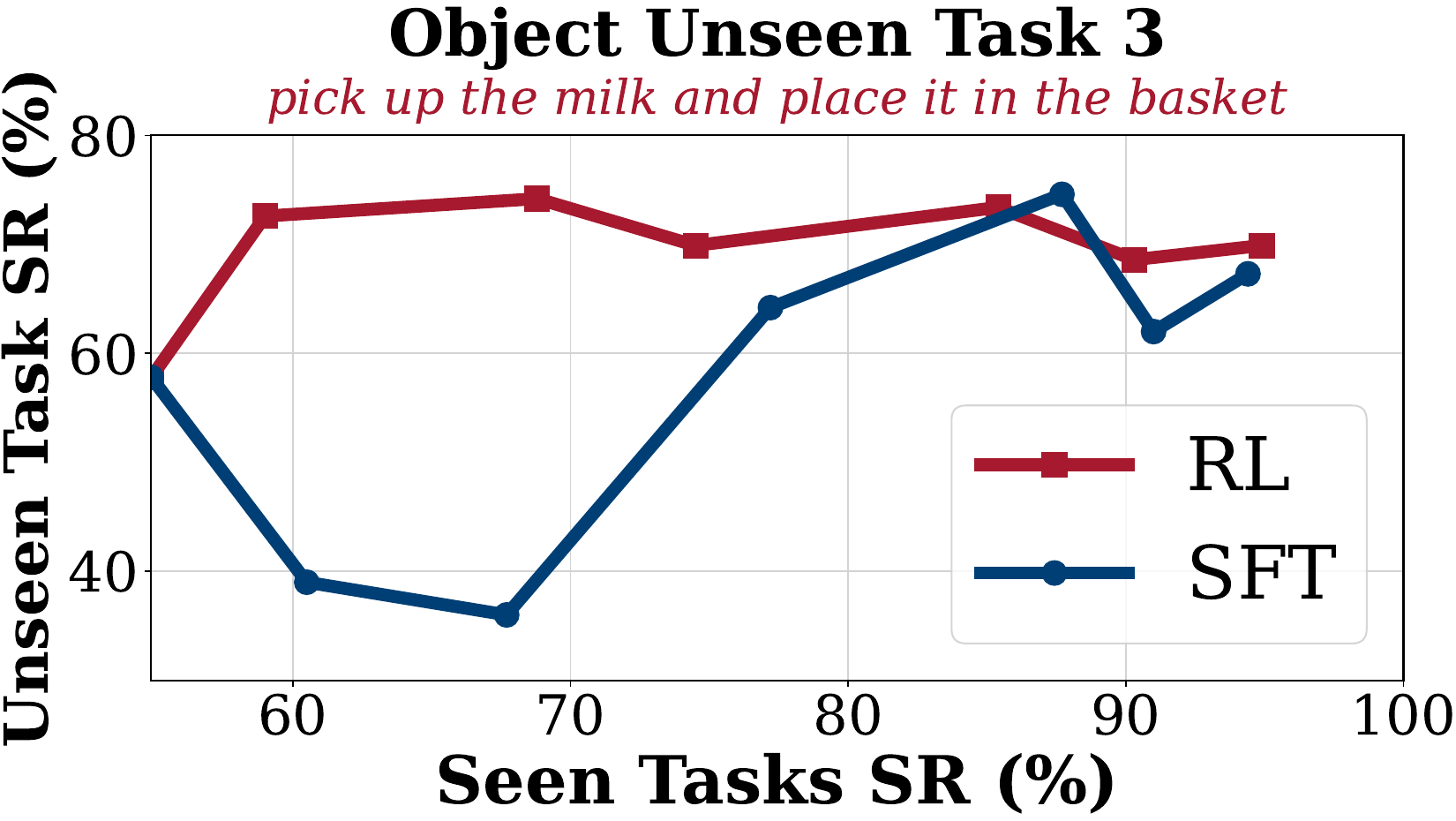}
        \label{fig:libero_object_c}
    \end{subfigure}

    \vspace{-0.2cm}

    \begin{subfigure}[b]{0.32\textwidth}
        \centering
        \includegraphics[width=\textwidth]{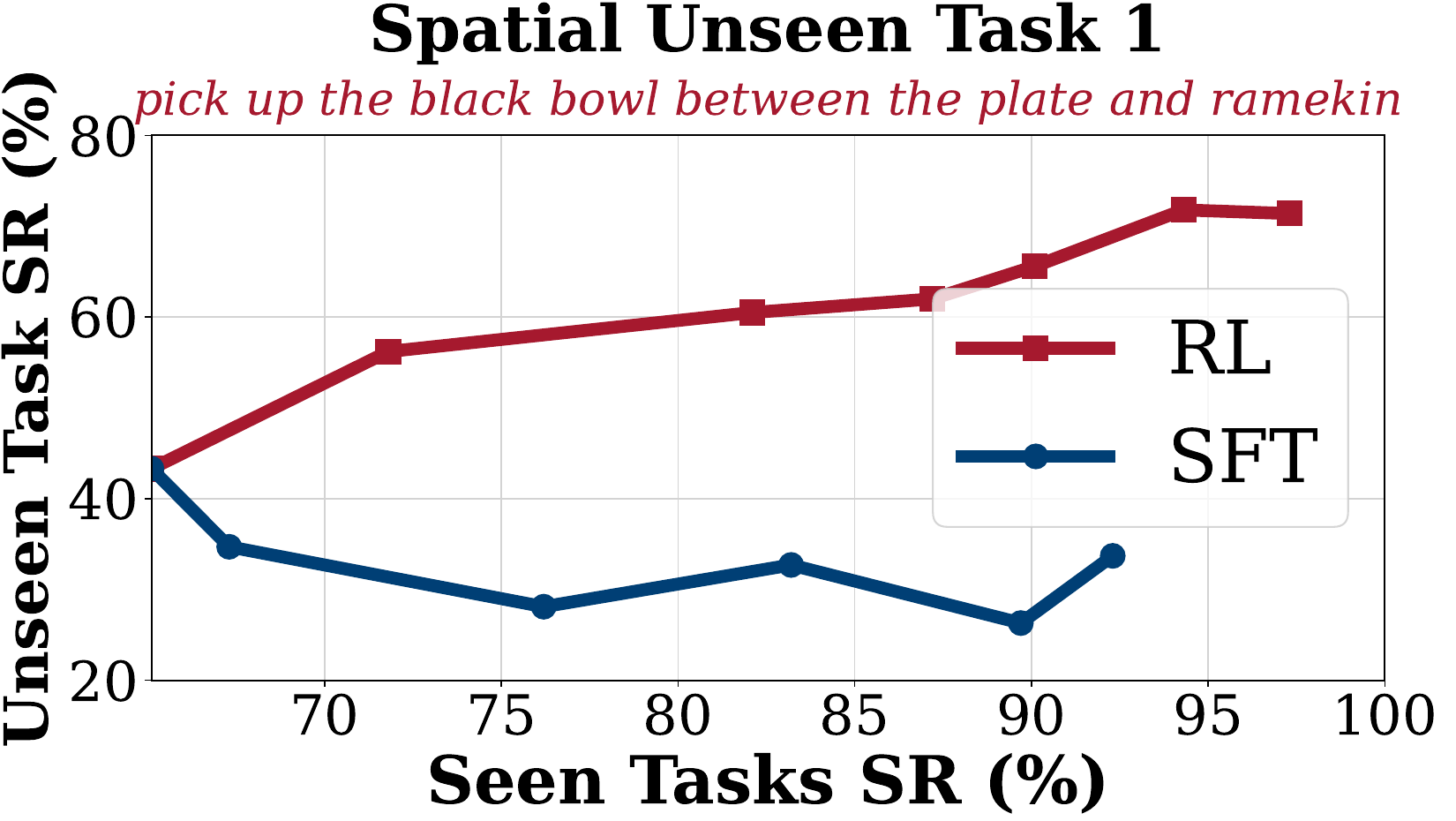}
        \label{fig:libero_spatial_a}
    \end{subfigure}
    \hfill
    \begin{subfigure}[b]{0.32\textwidth}
        \centering
        \includegraphics[width=\textwidth]{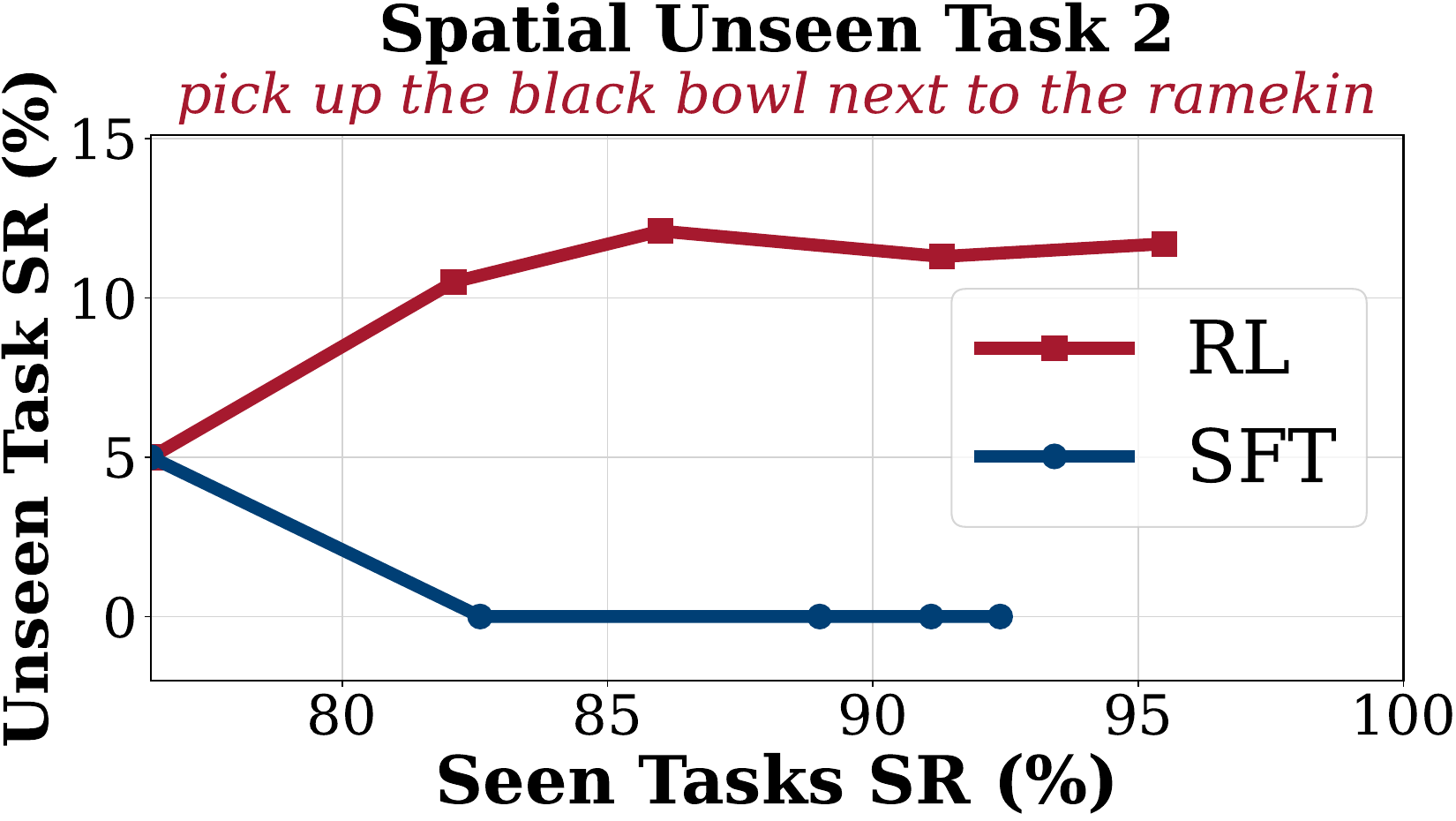}
        \label{fig:libero_spatial_b}
    \end{subfigure}
    \hfill
    \begin{subfigure}[b]{0.32\textwidth}
        \centering
        \includegraphics[width=\textwidth]{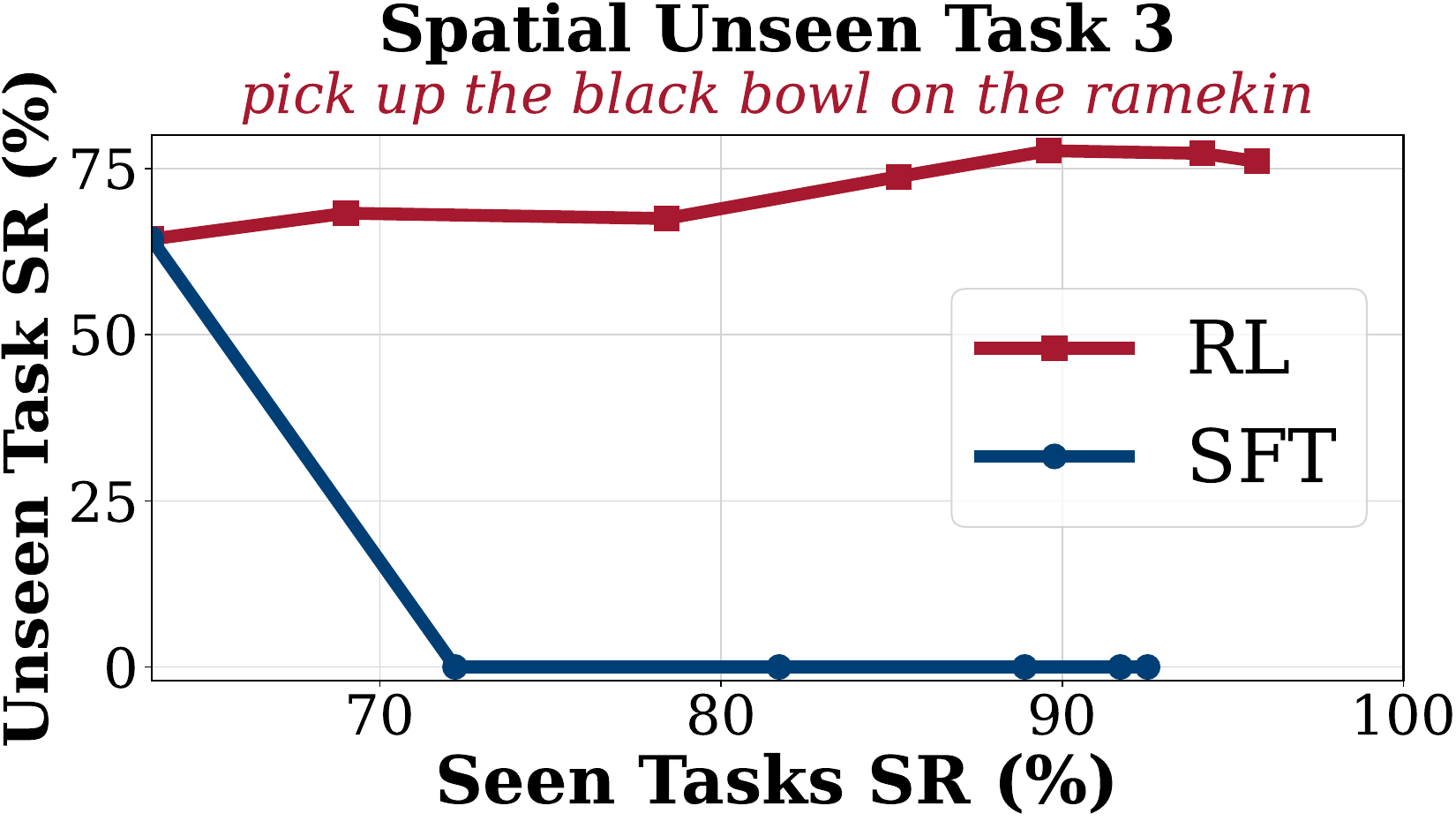}
        \label{fig:libero_spatial_c}
    \end{subfigure}

    \caption{Generalization Analysis on LIBERO: Goal Unseen~\textbf{(Top)}, Object Unseen~\textbf{(Middle)}, Spatial Unseen~\textbf{(Bottom)}.}
    \label{fig:libero_all_unseen}
    \vspace{-0.4cm}
\end{figure}

\textbf{Results\ }
Figure~\ref{fig:libero_all_unseen} presents the results.
During training, both SFT and RL achieve over 90\% success rates on training tasks.
However, their performance on unseen tasks diverges significantly.
As the seen tasks' success rates increase during training, \method shows consistent improvement on unseen tasks across different settings.
In contrast, SFT suffers from severe overfitting.
While achieving high performance on training tasks, SFT exhibits performance degradation on most unseen tasks, often shows catastrophic forgetting with success rates dropping to 0\%.
This indicates that RL training enables VLA models to retain previously acquired capabilities while learning generalizable skills from diverse tasks.
\textbf{On LIBERO-Goal's} three unseen tasks, the SFT success rates immediately drop to 0\% at the beginning of training.
This may be attributed to the fact that LIBERO-Goal tasks involve diverse objects and manipulation strategies with few transferable components across tasks, whereas SFT relies on similar task distributions to enable generalization.
In contrast, our \method avoids performance degradation and even achieves 5\%-15\% improvements on the three unseen tasks.
\textbf{On LIBERO-Object}, SFT improves from 57.8\% to 74.6\% on Unseen Task 3 but fails on the other two tasks.
In contrast, our method improves across all three unseen tasks, with gains of 36.5\% on Unseen Task 2 and 16.4\% on Unseen Task 3.
\textbf{On LIBERO-Spatial}, we observe a consistent trend: SFT degrades by 10\% on Unseen Task 1 and completely fails on the remaining tasks. In contrast, our method improves performance on Unseen Task 1 from 43.3\% to 71.8\% and achieves gains of 7.1\% and 13.3\% on the other two tasks.
\method consistently outperforms SFT in improving the generalization of VLA models, even in the most challenging cross-task scenarios.

\begin{table}[!t]
  \centering
  \caption{Real-world experiment~(sim2real) results.}
  \resizebox{.7\linewidth}{!}{
    \begin{tabular}{lccccc}
      \toprule
      & \textbf{Stack Bowls} & \textbf{Place Empty Cup} & \textbf{Pick Bottle} & \textbf{Click Bell} & \textbf{Avg} \\
      \midrule
      RDT & 60.0 & 4.0 & 10.0 & 20.0 & 23.5\\
      \midrule
      OpenVLA-OFT & 38.0 & 2.0 & 0.0 & 30.0 & 17.5 \\
       \textbf{w/ ours} & 70.0 & 10.0 & 14.0 & 60.0 & 38.5 \\
      \rowcolor{lightblue!100} \quad $\Delta$ & 
        \textcolor{red}{$+32.0$} & \textcolor{red}{$+8.0$}  & 
        \textcolor{red}{$+14.0$} & \textcolor{red}{$+30.0$}  &  \textcolor{red}{$+21.0$} \\
      \bottomrule
    \end{tabular}
  }
  \label{tab:sim2real}
\end{table}

\subsection{Real-World Experiments}\label{sec:real_world}

To evaluate the real-world effectiveness of \method, we conduct sim-to-real experiments on four RoboTwin2.0 tasks\footnote{The detailed descriptions of the four tasks can be found at: \url{https://robotwin-platform.github.io/doc/tasks/index.html}}: Stack Bowls, Handover Block, Pick Bottle, and Click Bell.
We employ OpenVLA-OFT as the policy model, RDT as the baseline model, and execute on two AgileX Piper robotic arms.
For each task, we first use 1000 simulation trajectories for SFT.
Then we apply~\method on the SFT model using 1000 simulation scenarios to obtain an RL model.
The entire training process uses only simulation data without any real-world demonstrations.
We evaluate both the SFT and RL models on clean tabletops with unseen backgrounds in the real world.
Each task is tested with 50 trials. The RDT baseline model only undergoes the SFT stage.

The sim2real results in Table~\ref{tab:sim2real} demonstrate that \method significantly improves the real-world success rates of VLA models, with an average improvement from 17.5\% to 38.5\%, surpassing RDT's 23.5\%. 
For instance, in the Stack Bowls task, \method achieves a 96\% relative improvement, lifting performance from 32\% to 70\% and outperforming RDT (60\%).
On the Pick Bottle task, which demands higher action precision, as the bottle will fall if the robotic arm is not perfectly aligned on the first attempt, the SFT model fails completely while \method achieves a 15\% success rate, demonstrating its effectiveness in improving action precision.
Using \method for low-cost, large-scale, and highly parallel RL training in simulation, we significantly improve the real-world performance of simulation-trained VLA models.
This demonstrates a promising path for scaling real-world policies: using rich simulation assets and high-fidelity simulators for cost-effective RL training to achieve superior performance in real-world deployment.

\section{Discussions}

\subsection{``Pushcut'': Emergence of New Patterns through RL}

\begin{figure}[!h]
    \centering
    \begin{minipage}[b]{0.495\textwidth}
        
        \centering
        \includegraphics[width=\linewidth]{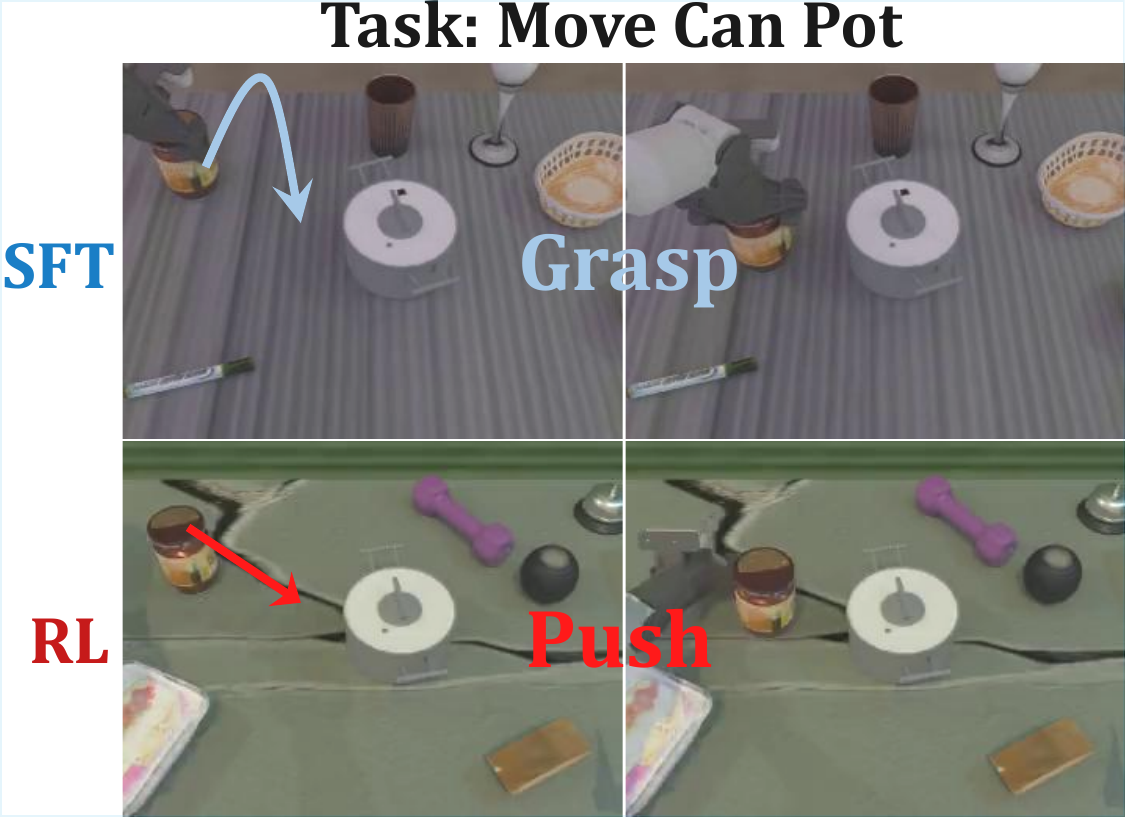}
        \subcaption{``move can pot'' task: Model learned to push the can to the pot (bottom) instead of grasp-move-place in the demonstration data (top).}
        \label{fig:move_can_pot}
    \end{minipage}
    \hfill
    \begin{minipage}[b]{0.495\textwidth}
        \centering
        \includegraphics[width=\linewidth]{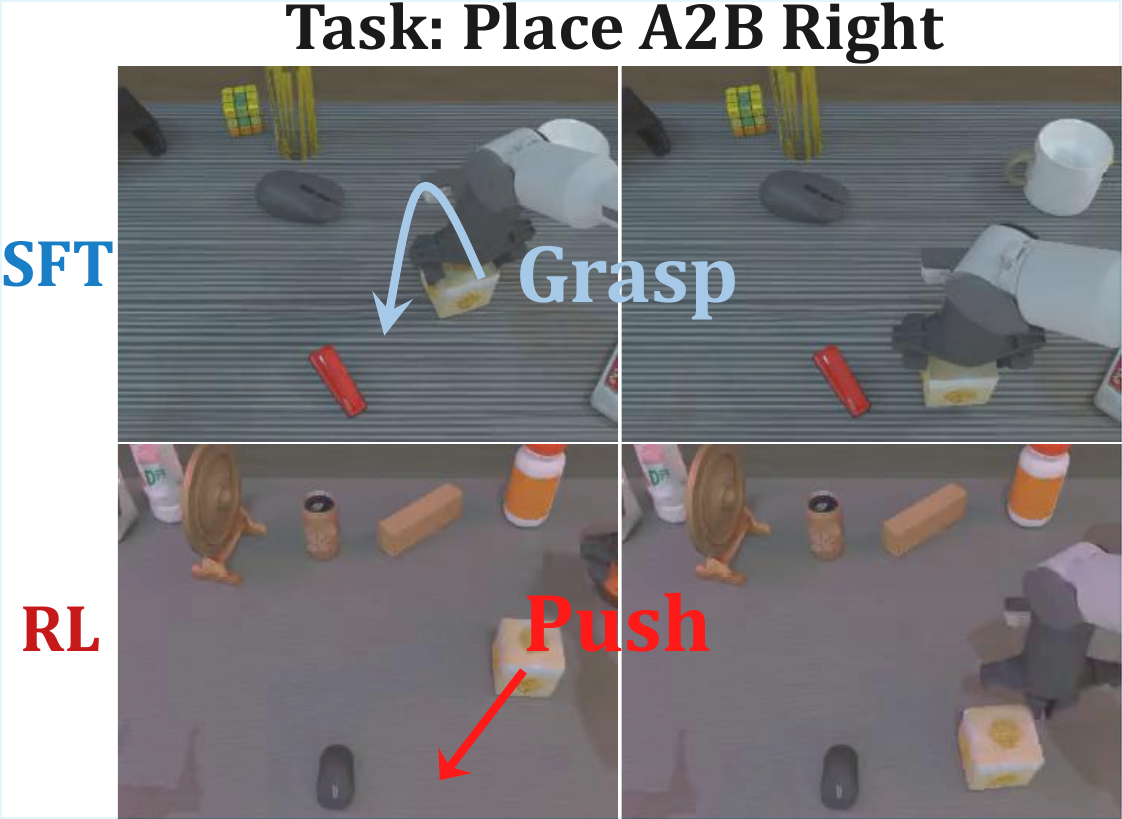}
        \subcaption{"place a2b right" task: Model learned to push A to B's right (bottom) instead of demonstrated grasp-move-place (top).}
        \label{fig:place_a2b}
    \end{minipage}
    \caption{Illustration of ``\textcolor{darkred}{pushcut}''. Emergent pushing behaviors through RL in RoboTwin2.0 tasks.}
    \label{fig:eight_images}
\end{figure}

In this subsection, we present an emergent behavior ``\textcolor{darkred}{pushcut}'' shown during RL training.
We observe that the VLA model learns novel behaviors that are absent from the demonstration data during the training process of~\method.
Specifically, in the \textit{\textbf{move can pot}} task of RoboTwin2.0, where the objective is to transport cans to positions adjacent to designated pots, all demonstration trajectories consistently adhere to a ``grasp–move–place'' strategy, as illustrated at the top of Figure~\ref{fig:move_can_pot}.

Remarkably, after RL training, the VLA model autonomously discovers a more efficient solution: instead of grasping, it pushes the can directly into the target location (Figure~\ref{fig:move_can_pot}, bottom). We refer to this emergent phenomenon as ``\textcolor{darkred}{pushcut}''~(a \textcolor{darkred}{push}-driven short\textcolor{darkred}{cut}), highlighting the model’s ability to bypass the conventional grasp–move–place routine.
Similar emergent behaviors are observed in the \textit{\textbf{place a2b left/right}} task.
While the demonstration data complete the task by grasping Object~A and placing it next to Object~B, the RL-trained model instead learns to accomplish the task by simply pushing Object~A into position, as shown in Figure~\ref{fig:place_a2b}.
The phenomenon of ``\textcolor{darkred}{pushcut}'' closely parallels the ``Aha Moment'' observed in DeepSeek-R1~\citep{guo2025deepseek}, as both emerge through RL, where the agent uncovers novel patterns.

This phenomenon highlights the fundamental distinction between SFT and RL. Whereas SFT merely replicates fixed patterns present in the demonstrations, RL facilitates exploration driven by rewards, thereby uncovering novel strategies.
During RL training, effective behaviors are reinforced through positive rewards, while less efficient ones are gradually eliminated.
The outcome-level reward further promotes such emergent strategies: since both grasping and pushing yield equivalent rewards upon successful completion, the sparse reward design avoids the procedural constraints of process-level supervision, affording the agent a broader exploration space and enabling the discovery of unanticipated yet effective solutions.

\subsection{Failure Modes of SimpleVLA-RL}
\label{sec:when_might_fail?}

\begin{table}[!htbp]
  \centering
  \caption{Impact of initial model capability on \method performance.}
  \resizebox{\linewidth}{!}{
    \begin{tabular}{lcccccc}
      \toprule
      \multirow{1}{*}{\textbf{}} &
        \multicolumn{6}{c}{\textbf{RoboTwin2.0}} \\
      \cmidrule(lr){2-7}
      & \textbf{Move Can Pot} & \textbf{Place A2B Lift} & \textbf{Place A2B Right} & \textbf{Place Phone Stand} & \textbf{Pick Dual Bottles} & \textbf{Avg} \\
      \midrule
      0 trajs SFT     & 0 & 0 & 0 & 0 & 0 & 0 \\
      \quad +RL       & 0 & 0 & 0 & 0 & 0 & 0 \\
      \midrule
      100 trajs SFT   & 9.4 & 7.8 & 7.8 & 10.1 & 1.2 & 7.3 \\
      \quad +RL       & 51.6 & 25.0 & 27.2 & 18.8 & 4.3 &  25.4 \\
      \rowcolor{lightblue!100} \quad $\Delta$ &
        \textcolor{red}{$+42.2$} &
        \textcolor{red}{$+17.2$} &
        \textcolor{red}{$+19.4$} &
        \textcolor{red}{$+8.7$} &
        \textcolor{red}{$+3.1$} &
        \textcolor{red}{$+18.1$} \\
      \midrule
      1000 trajs SFT  & 28.1 & 37.5 & 28.7 & 17.1 & 29.7 & 28.2 \\
      \quad +RL       & 61.2 & 45.3 & 37.5 & 39.6 & 68.3 & 50.4 \\
      \rowcolor{lightblue!100} \quad $\Delta$ &
        \textcolor{red}{$+33.1$} &
        \textcolor{red}{$+7.8$} &
        \textcolor{red}{$+8.8$} &
        \textcolor{red}{$+22.5$} &
        \textcolor{red}{$+38.6$} &
        \textcolor{red}{$+22.2$} \\
      \bottomrule
    \end{tabular}
  }
  \label{tab:base model}
\end{table}

This subsection investigates the failure conditions of \method and key influencing factors.
Through experiments on five RoboTwin2.0 tasks, we find that the model priors are the critical factor determining RL effectiveness.

\textbf{Settings\ }
Each task is trained under domain randomization with a single-task setting. We compare three model variants:
(1) the OpenVLA-OFT base model without trajectory fine-tuning (0 trajectories SFT);
(2) the model fine-tuned with 100 demonstration trajectories per task (100 trajectories SFT);
and (3) the model fine-tuned with 1000 demonstration trajectories per task(1000 trajectories SFT).
All models undergo \method training on 1000 training scenarios and are evaluated on 100 held-out test scenarios.

\textbf{RL fails completely when the base model has no initial task ability.}
Table \ref{tab:base model} reports the results. The base model (0-trajectory SFT) achieves a 0\% success rate across all tasks, exhibiting no task-relevant behaviors.
Despite extensive pretraining, OpenVLA shows extremely limited zero-shot generalization, consistent with findings in~\cite{kim2025fine}.
Because no successful trajectories are generated during sampling and only outcome rewards (without process rewards) are employed, every trajectory receives zero reward.
As a result, RL is unable to improve performance, which remains at 0\%.

\textbf{The model prior has a significant impact on the effectiveness of RL.}
Initial capability is strongly correlated with post-RL performance.
The 100-trajectory SFT model improves from 7.3\% to 25.4\% (an 18.1\% gain), while the 1000-trajectory SFT model improves from 28.2\% to 50.4\% (a 22.2\% gain) in average success rate.
This trend is consistent across tasks. For instance, in the \emph{move can pot} task, the 100-trajectory SFT model improves from 9.4\% to 51.6\%, whereas the 1000-trajectory SFT model improves from 28.1\% to 61.2\%.
These results highlight that stronger initial capabilities provide more effective starting points for exploration, thereby facilitating greater performance improvements.

\textbf{RL effectiveness has a threshold: when initial ability is too low, improvements remain negligible.}
Our findings further reveal that the effectiveness of RL is subject to a performance threshold.
When initial success rates are very low, online RL with outcome rewards yields only marginal improvements.
For example, in the \emph{pick dual bottles} task, the 100-trajectory SFT model improves from 1.2\% to 4.3\%, while the 1000-trajectory SFT model improves from 29.7\% to 68.3\%.
Similarly, in the \emph{place phone} task, the 100-trajectory SFT model gains 8.7\%, compared to a 22.5\% gain for the 1000-trajectory SFT model.
The results indicate that a minimal level of task competence is essential for effective RL. Below this threshold, exploration is ineffective and RL fails to produce meaningful gains.

\section{Related Works}

\subsection{Reinforcement Learning for Large Language Models}

Reinforcement Learning (RL) for Large Language Models (LLMs) has achieved remarkable success, demonstrating its ability to induce complex reasoning behaviors such as self-verification and iterative optimization, thereby significantly enhancing model performance on reasoning tasks~\citep{guo2025deepseek,jaech2024openai,liu2025understanding,cui2025process,zeng2025simplerl,zuo2025ttrl}.
Recent advancements in Large Reasoning Models (LRMs), such as DeepSeek-R1~\citep{guo2025deepseek}, highlight the effectiveness of RL in boosting reasoning capabilities even with simple rule-based rewards, as exemplified by GRPO~\citep{shao2024deepseekmath}.
This approach differs substantially from Reinforcement Learning from Human Feedback (RLHF)~\citep{ouyang2022training}, which aligns base models with human preferences using algorithms like Proximal Policy Optimization (PPO)~\citep{schulman2017proximal} and heavily relies on preference modeling.

Recent studies have increasingly focused on enhancing exploration in reinforcement learning to enable longer training horizons and improved performance.
DAPO~\citep{yu2025dapo} introduces Clip-Higher, a decoupled variant of PPO clipping, which sets a higher upper bound relative to the lower one (e.g., $\epsilon_\text{low}=0.2, \epsilon_\text{high}=0.28$). This adjustment allows low-likelihood but potentially valuable tokens to increase in probability, thereby encouraging exploration.
Building on this, POLARIS~\citep{Polaris2025} employs a staged curriculum of temperature increases (e.g., $0.7 \rightarrow 1.0 \rightarrow 1.1$ for a 7B model) to gradually expand trajectory diversity and facilitate more robust policy discovery.
In parallel, Entropy Mechanism~\citep{cui2025entropy} addresses entropy collapse, a persistent issue in extended training, through methods such as Clip-Cov and KL-Cov, which selectively clip probabilities or penalize high-covariance tokens to sustain effective exploration.
Similarly, ProRL~\citep{liu2025prorl} combines KL control with reference policy resetting to preserve stability and extend training without degrading performance.
A complementary line of work regulates entropy via temperature tuning. Acereason-nemotron 1.1~\citep{liu2025acereason} advocates adjusting temperatures to stabilize post-scaling entropy around a target (e.g., 0.3), balancing exploration and exploitation. \cite{liao2025enhancing} further proposes a dynamic scheduler that adapts temperature over time to maintain stable entropy, thereby supporting sustained performance gains.

\subsection{Vision Language Action Models}

In the field of robotic manipulation tasks, VLA models~\citep{kim2024openvla,kim2025fine,liu2024rdt,bu2025univla,hung2025nora,black2024pi_0,pertsch2025fast,intelligence2025pi_} have shown better performance and task generalization compared to traditional policy-based approaches~\citep{ma2022vip,yuan2024learning}.
These models integrate the VLM or LLM backbone with action modules through unified end-to-end training~\citep{zhong2025survey}.
This approach enables comprehensive multimodal understanding and fine-grained motor control~\citep{firoozi2025foundation}.
Currently, many studies are focused on enhancing the effectiveness of VLA models.
For example, E-COT~\citep{zawalski2024robotic,chen2025training} introduced Embedded Chain of Thought (ECoT) to improve the spatial reasoning ability of VLA models. RDT-1B and VPP~\citep{liu2024rdt,hu2024video} proposed diffusion-based frameworks for VLA models. Agibot world and Roboverse~\citep{geng2025roboverse,bu2025agibot} aim to build larger-scale simulation environments and trajectory datasets to improve the sim-to-real transfer and generalization capabilities of VLA models.
Additionally, Dexmimicgen~\citep{jiang2024dexmimicgen} explores automated methods to generate high-quality trajectory data to address the issue of data scarcity in robotics.
Despite the rapid advancements in the VLA domain, imitation learning remains the dominant training paradigm for VLA models~\citep{sapkota2025vision,kim2024openvla,kim2025fine,liu2024rdt,bu2025univla,hung2025nora,black2024pi_0,pertsch2025fast,intelligence2025pi_}.
Current VLA models typically follow a two-stage paradigm: pretraining on multimodal data (e.g., Open X-Embodiment~\citep{o2024open}) followed by SFT on collected robot trajectories. However, imitation learning is limited by its dependence on expensive, high-quality trajectory data and poor generalization to unseen scenarios.

\textbf{VLA RL Methods\ \ }
Recently, some efforts have attempted to apply RL to VLA training.
GRAPE~\citep{zhang2024grape} utilized Direct Preference Optimization (DPO)~\citep{rafailov2023direct} to train VLA models by integrating human preferences.
ConRFT~\citep{chen2025conrft} introduced Reinforced Fine-Tuning~\citep{trung2024reft} to train VLA models in real-world environments, iteratively training VLAs through alternating RL and SFT rounds.
ReinboT~\citep{zhang2025reinbot} focused on dense reward design and optimized VLA models through reward maximization.
\cite{guo2025improving} proposed an iterative training framework that combines Supervised Fine-Tuning (SFT) and RL stages to address training instability and computational overhead.
More recent works have further advanced VLA RL methods.
Our work is one of the earliest systematic explorations of VLA online RL, and we publicly released the code in May 2025\footnote{\url{https://github.com/PRIME-RL/SimpleVLA-RL}}.
Concurrently, RIPT-VLA~\citep{tan2025interactive} investigates a closely related problem, employing RLOO~\citep{ahmadian2024back} for VLA RL training.
Moreover, \cite{liu2025can} investigates RL's impact on VLA generalization capabilities, demonstrating significant improvements over SFT in unseen environments, objects, and textures.
RLinf~\citep{RLinf_repo} designed a flexible, scalable framework for VLA RL that unifies rendering, inference, and training, improving both VLA training efficiency and performance.
VLA-RL~\citep{lu2025vla} applies the PPO algorithm to the VLA model.
TGRPO~\citep{chen2025tgrpo} uses Claude3.7 to evaluate trajectories and optimizes VLA with GRPO.
RFTF~\citep{shu2025rftf} uses value models to generate dense rewards in embodied scenarios for VLA online RL.
Compared to the above works, our paper further explores the effectiveness of VLA RL on real-world robotic tasks.
We also conduct comprehensive analyses on how VLA RL addresses data scarcity challenges and improves policy generalization.

\section{Conclusion}

In this work, we present \method, an RL framework tailored for VLA models. By extending veRL with VLA-specific trajectory sampling and parallelized training–inference–rendering capabilities, \method enables scalable and sample-efficient online RL.
\method demonstrate significant improvements in data efficiency, generalization, and sim-to-real transfer.
The consistent performance gains across LIBERO and RoboTwin benchmarks highlight the potential of RL not only to alleviate the data scarcity challenge of SFT but also to substantially enhance the generalization capacity of VLA models.
We hope these findings pave the way for more autonomous and adaptable robotic models.

\bibliography{main}

\newpage

\appendix

\end{document}